\documentclass[11pt,a4paper]{article}
\usepackage{times,latexsym}
\usepackage{url}
\usepackage[T1]{fontenc}

\usepackage[acceptedWithA]{tacl2021v1}

\usepackage[utf8]{inputenc}
\usepackage{microtype}
\usepackage{inconsolata}
\usepackage{graphicx}
\usepackage{booktabs}
\usepackage{multirow}
\usepackage{array}
\usepackage{tcolorbox}
\usepackage{tabu}
\usepackage{xcolor,colortbl}
\usepackage{amsmath, amssymb, amsthm}
\usepackage{algpseudocodex}
\usepackage{mathabx}
\usepackage{pifont}
\usepackage{enumitem}
\usepackage[nameinlink]{cleveref}
\crefname{section}{\S}{\S\S}
\crefformat{section}{\S#2#1#3}
\usepackage{graphicx}
\usepackage{calc}
\newlength\myheight
\newlength\mydepth
\settototalheight\myheight{Xygp}
\def\ourFramework{\textsc{Co-FactChecker}}
\def\ourBaseFactchecker{\textit{Verifier}}

\newtheorem{theorem}{Theorem}
\newtheorem{corollary}{Corollary}[theorem]
\newtheorem{lemma}[theorem]{Lemma}
\newtheorem{assumption}{Assumption}

\usepackage{xspace,mfirstuc,tabulary}

\newif\iftaclinstructions
\taclinstructionsfalse 
\iftaclinstructions
\renewcommand{\confidential}{}
\renewcommand{\anonsubtext}{(No author info supplied here, for consistency with
TACL-submission anonymization requirements)}
\newcommand{\instr}
\fi

\iftaclpubformat 

\else

\fi

\title{\ourFramework{}: A Framework for Human-AI Collaborative Claim Verification Using Large Reasoning Models}

\author{
  \\
  \textbf{Dhruv Sahnan $^{1,*}$ \quad Subhabrata Dutta $^{2,*}$ \quad Tanmoy Chakraborty $^3$} \\
  \textbf{Preslav Nakov $^1$ \quad Iryna Gurevych $^2$} \\
  \\
  $^1$ MBZUAI, UAE \quad $^2$ TU Darmstadt, Germany \quad $^3$ IIT Delhi, India \\
  * Equal Contribution
}

\date{}

\begin{document}
\maketitle

\begin{abstract}
Professional fact-checkers rely on domain knowledge and deep contextual understanding to verify claims.
Large language models~(LLMs) and large reasoning models (LRMs) lack such grounding and primarily reason from available evidence alone, creating a mismatch between expert-led and fully automated claim verification.
To mitigate this gap, we posit human--AI collaboration as a more promising path forward, where expert feedback, grounded in real-world knowledge and domain expertise, guides the model's reasoning. However, existing LRMs are hard to calibrate to natural language feedback, particularly in a multi-turn interaction setup.
We propose \textbf{\ourFramework{}}, a framework for human--AI collaborative claim verification.
We introduce a new interaction paradigm that treats the model's thinking trace as a shared scratchpad.
\ourFramework{} translates expert feedback into \textit{trace-edits} that introduce targeted modifications to the trace, sidestepping the shortcomings of dialogue-based interaction.
We provide theoretical results showing that trace-editing offers advantages over multi-turn dialogue, and our automatic evaluations demonstrate that \ourFramework{} outperforms existing autonomous and human--AI collaboration approaches.
Human evaluations further show that \ourFramework{} is preferred over multi-turn dialogue, producing higher quality reasoning and verdicts along with relatively easier to interpret and more useful thinking traces.\footnote{Code released on \href{https://github.com/UKPLab/arxiv2026-co-fact-checker.git}{GitHub}.}
\end{abstract}

\section{Introduction}
\label{sec:introduction}
Large language models (LLMs) and large reasoning models (LRMs) have fueled interest in automating time-intensive tasks to reduce expert workload.

\noindent However, these models fall short on complex tasks where accurate decision-making requires deep contextual understanding and domain knowledge, such as fact-checking.

\paragraph{Fact-checking requires expert knowledge.}
Accurate verification of real-world claims is a fundamentally complex expert task, in part because many misinformation claims are neither true nor false; rather, they blend elements of both ~\cite{nakov_assistingfcs,warren_2025_showmethework}.
Debunking such a claim using a clear veracity label is hard, especially when the label space is fine-grained and some labels are closely related, which is common in the label sets used by many fact-checking organizations (e.g., Snopes uses labels like mostly true, mixture and mostly false; while USA Today uses labels like partly false, false implication, and missing context)~\cite{juneja_mitra_humanFC_2022,snopes_fcratings,ustoday_fcratings}.
Disambiguating between such close labels requires contextual knowledge, much of which is embedded in the broader narrative surrounding the claim and is not always apparent from the evidence alone.
Moreover, a critical part of claim verification is to lay out clear, logical, and evidence-based reasoning steps to explain the claim~\cite{graves_epistemology_2017,das_humancenterednlpFC_2023}, as well as to decide \textit{which} evidence and the corresponding reasoning steps matter the most, \textit{why} they matter, and \textit{how} they should be weighed relative to each other.
This reasoning must clearly place the claim into proper context against the broader narrative and fully convince any reader of the verdict~\cite{Lewandowsky2020DebunkingHandbook}.

\paragraph{A case for Human--AI collaboration.} Fact-checkers draw on ethical judgment, intuition, and human values --- often shaped by experience and cultural, social, and domain knowledge --- to identify and reason from relevant evidence to verify a claim~\cite{micallef_truefalse_2022,warren_2025_showmethework}.

\begin{figure*}[t]
    \centering
    \includegraphics[width=0.9\linewidth]{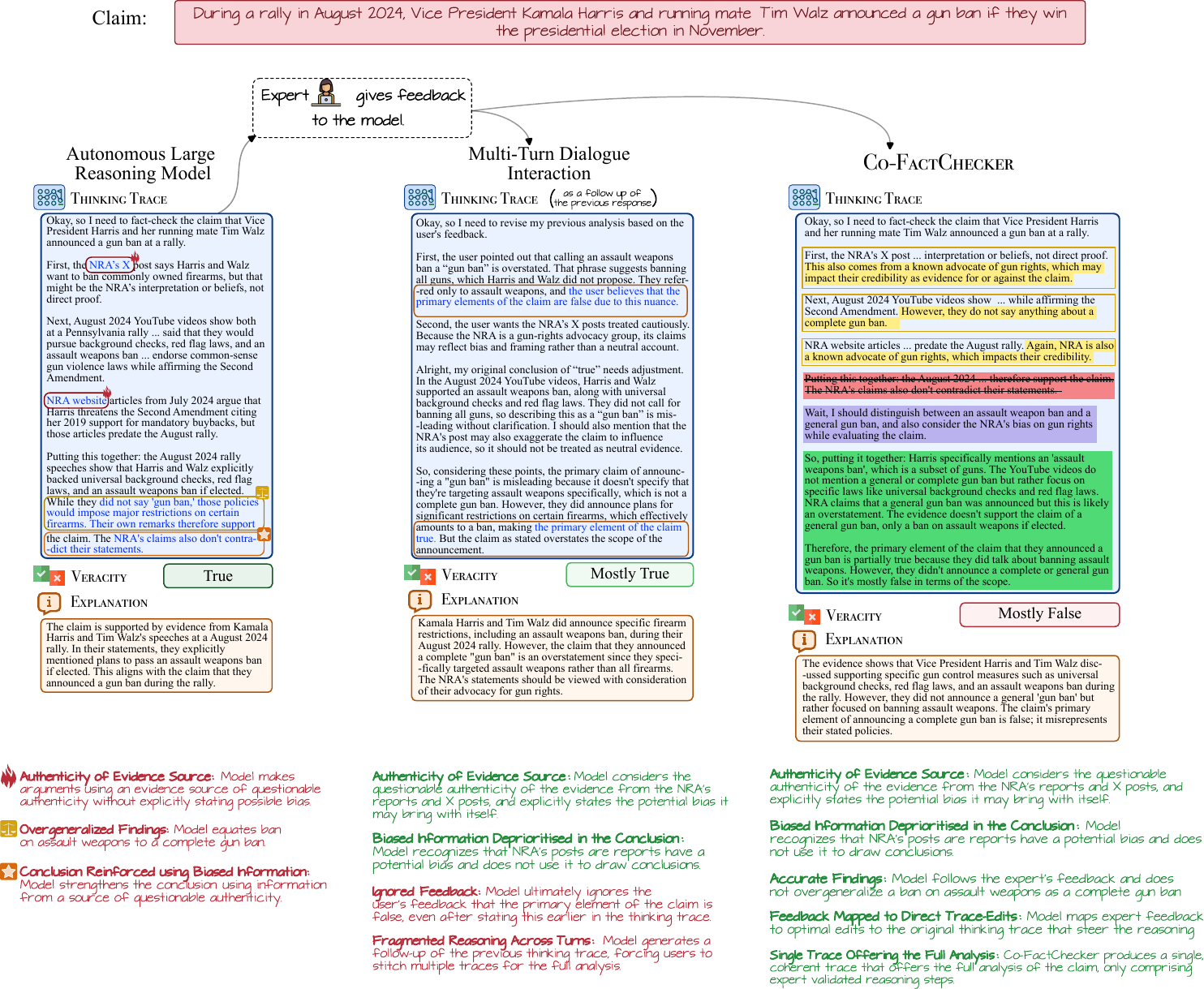}
    \caption{\textcolor{black}{\textbf{Sample run of the claim verification task for a given claim.} We show the response of the autonomous LRM, for which the expert provides some feedback. We show the response of the feedback integration in two interaction setups: multi-turn dialogue interaction and \ourFramework{}. We see that the autonomous model makes some errors in the reasoning; the model is unable to integrate all feedback points into the reasoning in multi-turn dialogue; and \ourFramework{} resolves these failure modes through its trace-editing mechanism.}}
    \label{fig:intro_fig}
\end{figure*}

\newpage
Autonomous LLM-based systems lack such background knowledge and typically rely only on the available evidence and their parametric knowledge.
\textcolor{black}{For instance, as illustrated in \Cref{fig:intro_fig}, the autonomous LRM constructs arguments using evidence from an inauthentic source and uses these arguments to draw conclusions while verifying a given claim. 
It also overgeneralizes the core finding and incorrectly concludes that the claim is true.
In contrast, professional fact-checkers did not use the inauthentic source as evidence, rather as an example of where the claim was found to be spreading; and correctly contextualized the claim by distinguishing between ``an assault weapons ban'' and ``a complete gun ban''.}
Moreover, fact-checkers have also repeatedly expressed their skepticism towards tools that aim to fully automate claim verification, arguing that human expertise cannot be replaced~\cite{fullfact_2020,nakov_assistingfcs,warren_2025_showmethework}.

Instead, fact-checkers have advocated for hybrid, human-in-the-loop approaches as the way forward~\cite{das_humancenterednlpFC_2023}.
A natural paradigm for human--AI collaboration in this context would be as follows: an LLM/LRM proposes a verdict comprising a veracity label and a short fact-checking explanation along with an explicit decision-making process; a human fact-checker interacts with the model by validating the model's reasoning and the verdict; and provides feedback, embedded with contextual knowledge, which is used as guidance to iteratively co-construct a more accurate verdict.

\paragraph{Challenges of effective collaboration.} However, designing a framework that enables effective human--AI collaboration is non-trivial.
Even though \textit{multi-turn dialogue} emerges as a natural interaction protocol, recent studies reveal issues that may undermine human--AI collaboration for verifying claims.

First, existing automatic fact-checking systems often ``black-box'' their decision-making process, making it difficult for users to understand the rationale behind the proposed verdicts~\cite{micallef_truefalse_2022,das_humancenterednlpFC_2023}.
Explicit reasoning in LRMs mitigates this issue to an extent, but gaining control over the reasoning trajectories of these models remains difficult through prompting alone~\cite{kwon2025reasoniflargereasoningmodels}.
LRMs frequently struggle to incorporate user input, especially when interacting in multi-turn dialogue: ignoring parts of the input, failing to retain global instructions or such given in preceding interactions, and errors propagating due to a previously proposed problematic responses~\cite{li2025thinkingfailspitfallsreasoning,kwan-etal-2024-mt-custom}.
In expert domains, these limitations only become more pronounced~\citep{sahinuc2025expertpreferencebasedevaluationautomated}.
\textcolor{black}{Consistent with these findings from existing work, for the claim verification task, we observed that the LRM often failed to integrate expert feedback into the reasoning and failed to retain global instructions about the claim verification task in a multi-turn dialogue (c.f. \Cref{fig:intro_fig} and \Cref{sec:app:intro_egs}).
Moreover, this setup also produced reasoning that was fragmented across several turns, which forced users to stitch together multiple thinking traces to reconstruct the full reasoning (c.f. \Cref{sec:human_evals}).}

\paragraph{Proposed solution.} To address these challenges, we introduce \textbf{\ourFramework{}}, a framework for human-AI collaborative claim verification.
Our framework uses an LRM that first proposes a candidate verdict while exposing its decision-making process through a \textit{thinking trace}.
This trace serves as a common scratchpad between the fact-checker and the LRM to co-construct the reasoning.
Here, the fact-checker validates the reasoning steps in the trace and instructs the framework to add, to remove, or to modify them using natural language feedback.
To support this interaction, our framework introduces an editor module that is designed to interpret the fact-checker's feedback and to apply optimal edits to the thinking trace, and subsequently to derive the verdict for the claim.

Our main contributions are four-fold:

\begin{itemize}[label=\ding{229}]
    \item  We propose \textbf{\ourFramework}, a framework for claim verification, which introduces a new interaction paradigm that enables a fact-checker and an LRM to reason and to co-construct evidence-based reasoning steps within a shared thinking trace, and subsequently to derive a verdict for the input claim.

    \item We provide theoretical proofs showing that trace-editing can dominate multi-turn dialogue under equal capacity constraints and realistic assumptions aligned with real LRMs.

    \item Through comprehensive automatic evaluation, we show that \ourFramework{} outperforms existing interaction approaches that facilitate human--AI collaboration as well as autonomous LLM-based systems.

    \item Finally, through human evaluations, we show that \ourFramework{} is preferred over multi-turn dialogue, offering better control to users for guiding the model, better reasoning and verdicts, and thinking traces that are relatively easier to perceive and more useful.
    We also discuss key issues as experienced by users to facilitate future research.
\end{itemize}

\section{Related Work}
\label{sec:background}

\paragraph{Automatic fact-checking.}
Prior work has largely pursued fully automating fact-checking by splitting the problem into several subtasks~\cite{vlachos-riedel-2014-fact,guo-etal-2022-survey,sahnan2025llmsautomatefactcheckingarticle}.
However, human-centered analyses found that fact-checking experts see limited utility in autonomous tools~\cite{juneja_mitra_humanFC_2022,das_humancenterednlpFC_2023}, and remain skeptical, especially of automatic \textit{claim verification}, where human intuition, creativity and expert judgment are critical, yet not reliably captured by current models~\cite{nakov_assistingfcs}.
Many claims are not binary; rather they mix true and false elements, omit crucial context, or are true only under specific framings~\cite{nakov_assistingfcs,warren_2025_showmethework}. 
Accordingly, fact-checking organizations use fine-grained veracity labels paired with short explanations \cite{snopes_fcratings,ustoday_fcratings}.
Fact-checkers rely on domain knowledge and expertise to place such claims into proper context.
Moreover, fact-checkers do not just \textit{cite} evidence; they construct evidence-based reasoning to explain the claim and weigh the importance and 
the reliability of each piece of evidence against the broader real-world context to derive a verdict \cite{graves_epistemology_2017,warren_2025_showmethework}.
In contrast, LLMs/LRMs may lean on surface-level associations and statistical priors~\cite{loru_simulationofjudgement_2025}, and existing claim verification systems often ``black-box'' their decision-making process~\cite{das_humancenterednlpFC_2023}.
These observations motivate two design choices for our work.

To better align with real-world practices, we (\emph{i})~treat organization-specific \textit{veracity labels} as part of the input, and (\emph{ii})~expose the model's decision-making process via thinking traces to support expert oversight.

\paragraph{Human-AI collaboration.} 
Instead of optimizing for full automation, recent work argues for human--AI collaboration on complex, expert-domain tasks, where humans and models co-construct solutions that align with expert knowledge and preferences \cite{dutta_haico_2025,lou2025unravelinghumanaiteamingreview,zou2025llmbasedhumanagentcollaborationinteraction}.
Prior studies operationalize collaboration via natural language interaction, e.g., multi-turn dialogue for problem-solving~\cite{lin-etal-2024-decision,shao2025collaborativegymframeworkenabling}, preference-based search over the solution space~\cite{dutta_haico_2025}, and multi-turn rewards to train models to collaborate \cite{wu2025collabllm}.
Claim verification is one such expert-domain task, and fact-checking experts have advocated for human-in-the-loop approaches~\cite{das_humancenterednlpFC_2023}.
Motivated by these insights, we propose \ourFramework{}, enabling experts to collaborate with an LRM using natural language feedback, and to co-construct evidence-based reasoning steps to derive the verdict for a given claim.
Our approach builds on the preference-based search paradigm of \citet{dutta_haico_2025}, in which we map expert feedback to targeted edits on the model's decision-making process to derive a more accurate verdict. 

\paragraph{Intervening in the thinking trace.}
Towards mitigating the lack of instruction-following and controllability in LRMs   \cite{li2025thinkingfailspitfallsreasoning,kwan-etal-2024-mt-custom,laban2025llms, kwon2025reasoniflargereasoningmodels}, recent efforts intervene on intermediate \textit{thoughts} to steer model behavior.
For example, \citet{li2025thinkpilotsteeringreasoningmodels} add task-specific think-prefixes, and \citet{zhang-etal-2025-steer} insert targeted thoughts at selected positions within a trace.
Our approach goes beyond such insertions by allowing explicit modification and removal of existing reasoning steps, enabling finer-grained correction of erroneous reasoning and more direct control over reasoning trajectories.
More importantly, prior work has not explored trace-editing as an \emph{interaction paradigm} between humans and models.
In our framework, we treat the thinking trace as a shared scratchpad: expert feedback is translated into direct edits, and the revised trace is used to continue generation and to derive a more accurate verdict.

\begin{figure*}[t]
    \centering
    \includegraphics[width=0.68\linewidth]{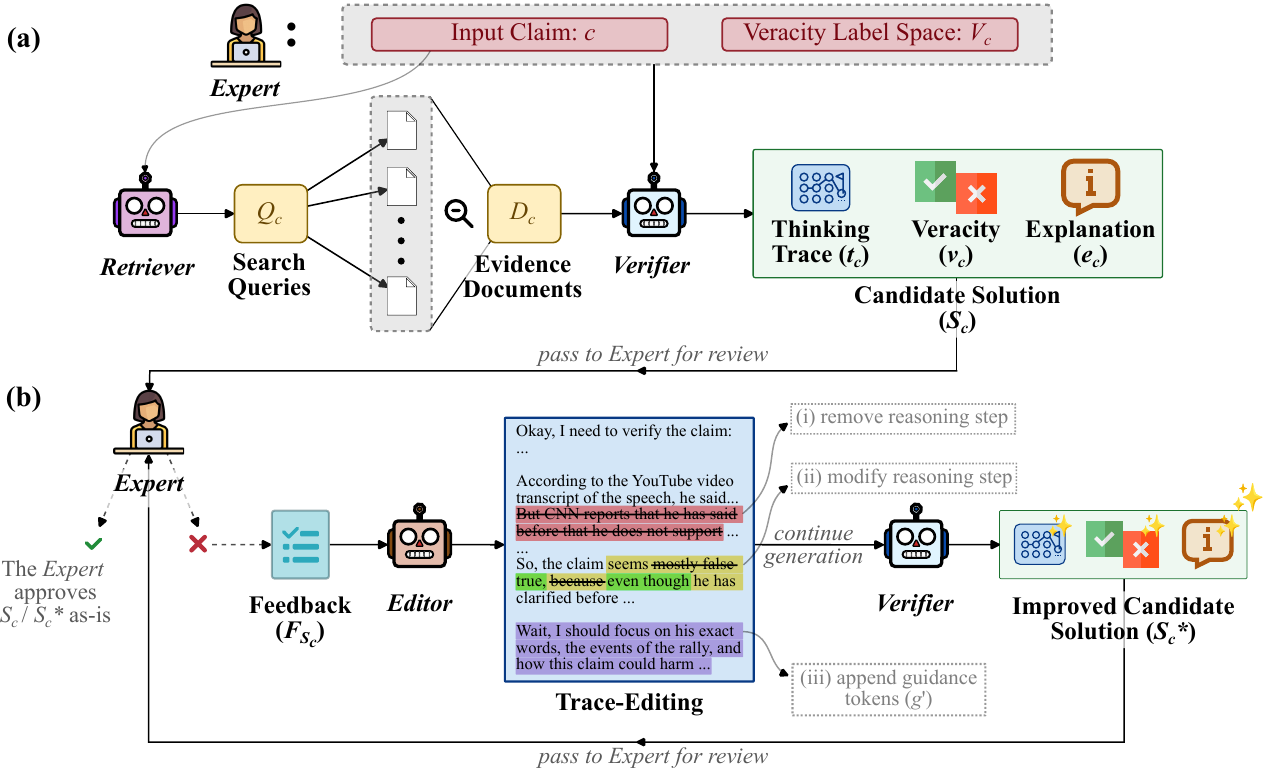}
    \caption{\textbf{Workflow of \ourFramework{}.} We split the collaboration process into two stages: (a) \textit{proposing a candidate solution}: the \textit{Retriever} retrieves evidence and the \textit{Verifier} produces a verdict for the claim, and a thinking trace; (b) \textit{co-constructing an improved solution}: the \textit{Expert} reviews the solution, the \textit{Editor} converts expert feedback into trace-edits, and the \textit{Verifier} continues the generation from the edited trace to yield a refined verdict and trace.}
    \label{fig:our_method}
\end{figure*}
\section{Method}
\label{sec:method}
\subsection{Task Overview}
Given a claim $c$ and a user-specified veracity label
set $V_c$, produce (\textit{i}) a verdict for $c$, comprising
a veracity label $v_c \in V_c$ and a short fact-checking explanation $e_c$; and (\textit{ii}) a thinking trace $t_c$ that documents the model's decision-making \linebreak process.
We denote a candidate solution $S_c$ as:
\begin{equation*}
    S_c = \langle v_c, e_c, t_c \rangle,
\end{equation*}
and we represent the thinking trace $t_c$ as an ordered list of evidence-based reasoning steps $r_i$:
\begin{equation*}
    t_c = (r_1, r_2, \ldots, r_m).
\end{equation*}

We further study a human--AI collaborative setup for this task, in which expert knowledge $k$ is injected into the model.
This knowledge may be injected implicitly (e.g.,~the expert selects one among $N$ solutions) or explicitly (e.g.,~the expert reviews a solution and gives natural language feedback to guide the model towards an improved solution).

\subsection{\ourFramework{}}
We propose \ourFramework{}, a framework for human--AI collaborative claim verification.
We design our framework around the key idea of treating an LRM's thinking trace as a \textit{shared scratchpad}: the model proposes a verdict while exposing its reasoning steps, the expert reviews them, and the framework converts the expert feedback into direct trace-edits that steer the reasoning.
\ourFramework{} comprises two main components:

\paragraph{{\bf\ding{229}}  The \textit{Verification Modules}} propose a candidate solution for the claim verification task and integrate expert feedback to improve the solution. This component comprises three modules:

(\textit{i}) \textit{Retriever} $\boldsymbol{\mathcal{R}}$, the evidence retriever, which generates search queries and retrieves relevant evidence from a search database.

({\textit{ii}) \textit{Verifier}\footnote{Not to be confused with verifier modules in test-time scaling literature~\cite{lightman2024lets, lifshitz2025multiagentverificationscalingtesttime}.} $\boldsymbol{\mathcal{V}}$, the base LRM, which reasons over retrieved evidence to draft a candidate solution for the claim verification task.

(\textit{iii}) \textit{Editor} $\boldsymbol{\mathcal{E}}$, which maps expert feedback into trace-edits that remove or modify reasoning steps, or guide the LRM along a specific reasoning path.
We pair this module with a custom reward model (cf. \Cref{sec:app:editor_rm}), which allows us to sample better actions for performing trace-edits.

\paragraph{{\bf\ding{229}} The \textit{Expert}} reviews a candidate solution and provides natural language feedback, grounded in expert knowledge, to highlight and to improve problematic reasoning steps in $\boldsymbol{\mathcal{V}}$'s thinking trace.

As illustrated in \Cref{fig:our_method}, we split the collaborative verification process into two stages as follows:
\paragraph{(a) Proposing a candidate solution}\ 
\newline
Given claim $c$, the retriever $\boldsymbol{\mathcal{R}}$ generates a set of search queries $Q_{c}$ to extract more context about the claim and its surrounding narrative.
For each query $q \in Q_{c}$, it retrieves the top-$k$ documents from the search database, yielding a set of relevant evidence documents $D_c$.
Now, the verifier $\boldsymbol{\mathcal{V}}$ uses $D_c$ and $V_c$ to verify the claim $c$, producing a verdict comprising a veracity label $v_c \in V_c$ and a fact-checking explanation $e_c$, along with a thinking trace $t_c$.
Together, they form the candidate solution $S_c$:
\begin{align*}
    S_{c} &=\ \boldsymbol{\mathcal{V}}(c, E_{c}, V_{c})\\ 
    &=\ \langle v_{c}, e_{c}, t_{c}\rangle
\end{align*}

\paragraph{(b) Co-constructing an improved solution}\ 

\paragraph{The \textit{Expert} gives feedback.}
Here, the expert reviews the proposed solution $S_{c}$, primarily verifying whether the reasoning steps in $t_c$ constructed using $D_c$ are correct; and whether $v_c$ and $e_c$ correctly place the claim $c$ into context.
Based on their review, the expert either accepts $S_c$ as-is, or returns a set of natural language feedback instructions $F_{S_c}$, grounded in expert knowledge $k$.

We denote expert feedback step as follows:
\begin{align*}
    F_{S_{c}} &=\ \text{ExpertReview}(S_{c}, k) \\
    &=\ 
    \begin{cases}
        \emptyset & ;\ S_{c}\text{ accepted} \\
        \{f_1, f_2, \ldots, f_n \} & ; \text{ otherwise},
    \end{cases}
\end{align*}
where each $f_i$ represents a feedback instruction, which identifies an issue in $S_c$ (e.g., an incorrect reasoning step, speculative framing, or missing real-world context).

Moreover, to enable automatic evaluations at scale, we design an \textbf{LLM-based \textit{oracle}} $\boldsymbol{\mathcal{O}}$ to act as a proxy for the human expert.
As discussed, experts rely on important context from real-world experience and domain knowledge (which shape their expert knowledge $k$) for the claim verification task.
LLMs, however, lack such context, which limits their ability to validate candidate solutions just based on their internal parametric knowledge.
We therefore allow $\boldsymbol{\mathcal{O}}$ to refer to the ground-truth fact-check and to fine-grained grading rubrics (cf. \Cref{sec:app:oracle_rubrics}), which together form synthetic expert knowledge $k^{syn}$.
Now, the oracle produces an evaluation report $\texttt{Report}_{S_c}$ for $S_c$ using $k^{syn}$ and then converts it into feedback instructions $F_{S_c}^{\mathcal{O}}$:
\begin{align*}
\texttt{Report}_{S_c} &= \boldsymbol{\mathcal{O}}(S_c, k^{syn}),\\
F^{\mathcal{O}}_{S_c} &= \boldsymbol{\mathcal{O}}(S_c, \texttt{Report}_{S_c}),
\end{align*}
where $F^{\mathcal{O}}_{S_c}$ follows the same format as the expert feedback and can drive trace-edits at scale.

\paragraph{Trace-editing.}
Here, the editor $\boldsymbol{\mathcal{E}}$ translates the expert feedback $F_{S_c}$ into targeted trace-edits over the original trace $t_c=(r_1,\ldots,r_m)$.
In particular, $\boldsymbol{\mathcal{E}}$ first assigns each instruction $f_i \in F_{S_c}$ to one of three trace-edit types:
(\textit{a})~remove a reasoning step $r_i$ from $t_c$,
(\textit{b})~modify a reasoning step $r_i \in t_c$ to a correction $r_i'$,
and (\textit{c})~guide global reasoning by appending guidance tokens $g'$ to $t_c$.
$\boldsymbol{\mathcal{E}}$ then iterates over each $r_i \in t_c$ and applies all remove and modify trace-edits, yielding the edited trace $t_c'$.
Finally, it compresses all guide-type instructions into a single, compact sequence of guidance tokens $g'$ using a predefined template, so that guidance is injected in a consistent, model-readable form.

\paragraph{Continuing the generation for a new solution.}
Now, $\boldsymbol{\mathcal{V}}$ \emph{continues the generation} using the edited trace $t_
c'$, appended with guidance tokens $g'$ that nudge $\boldsymbol{\mathcal{V}}$'s reasoning trajectory along a specific path,  as a prefix (suppressing the previously generated verdict and the end-of-thinking token), producing an improved candidate solution $S_c^*$.
Formally, $\boldsymbol{\mathcal{V}}$ extends the prefix to a new trace $t_c^*$ and produces an updated verdict comprising a new veracity label $v_c^* \in V_c$ and a new explanation $e_c^*$, as follows:
\begin{equation*}
S_c^* = \langle v_c^*,e_c^*, t_c^* \rangle .
\end{equation*}
Since the generation continues from the edited prefix, the updated trace has the form
\begin{equation*}
t_c^* = t'_c + g' + (r_1^*, r_2^*, \ldots, r_{m'}^*),
\end{equation*}
where $(r_1^*, \ldots, r_{m'}^*)$ are new reasoning steps
produced after integrating the expert feedback.
We continue the \textit{solution} $\rightarrow$ \textit{feedback} $\rightarrow$ \textit{trace-edits} $\rightarrow$ \textit{solution} loop until the expert is satisfied with the solution.
For automatic evaluations, we fix the maximum number of loop iterations to 3 (cf. \Cref{sec:app:cofc_details} for additional details).

\section{Theoretical Results}
\label{sec:theory}
Here, we provide theoretical guarantees for why our trace-editing approach is provably better to integrate natural language feedback into a solution compared to a dialogue setup.

Let $\Sigma$ be a finite vocabulary of tokens. An LRM generates a thinking trace $t\in \Sigma^{*}$ followed by a decision $y$. The human expert observes a discrete random correction variable $C$ with support $\cal C$.
Let ${\cal L}(t)=\mathbb{E}\left [ l(y, y^{*})|t\right ]$ denote the Bayes risk induced by trace $t$, where $l(\cdot)$ denotes a loss function over decision $y$ inferred from trace $t$ given the ground truth decision $y^{*}$.

We assume the loss depends on a trace only through its risk-relevant content: there exists a projection $\pi:\Sigma^{*}\!\rightarrow\!\mathcal{W}$ (the argumentatively load-bearing facts and steps) such that ${\cal L}(t)$ depends on $t$ only through $\pi(t)$. This matches the claim verification setting, where only the argumentative relationship between facts is relevant.
We denote the expected risks induced by the trace-edit and the dialogue channel as $L_\tau$ and $L_{F_{<y,t>}}$, respectively. $I(X;Y)$ denotes mutual information between random variables $X$ and $Y$.
We consider the following two channels of oracle interventions.

\noindent{\bf\ding{229} Trace-edit (non-monotone).} There exists a finite set of traces $\tau (t)\subset \Sigma^{*}$ reachable from the current trace $t$ via a finite set of edits, with $|\tau (t)|=M_\text{edit}$. An intervention is specified by an \emph{edit program} $A\in\mathcal{A}$ that may \emph{delete or overwrite} a committed span of $t$; the resulting trace $t'=\mathrm{Apply}(t,A)\in\tau(t)$ \emph{replaces} $t$ as the state from which subsequent reasoning proceeds.

\noindent{\bf\ding{229} Dialogue (append-only).} The oracle emits textual feedback (possibly multi-turn) $F_{<y,t>}$. The model maps this message to a new trace $t_{F_{<y,t>}}$ via an arbitrary function of $(t, F_{<y,t>})$. Crucially, the erroneous trace $t$ \emph{persists in the conditioning context}---the effective state is $(t, y, F_{<y,t>})$---so feedback can only be \emph{added}, never removed.


\textcolor{black}{In either channel, we write $s$ for the resulting \emph{conditioning state}, i.e., the context on which the decision is drawn, and extend the risk to states via $\mathcal{L}(s)=\mathbb{E}\left[l(y,y^{*})\mid s\right]$, so that $\mathcal{L}(t)$ is the special case $s=t$. Concretely, $s=(t,y,F_{<y,t>})$ for the dialogue and $s=t'$ for the trace-edit channel.}
\begin{assumption}[Symmetric bottleneck]
\label{as:bottleneck}
The model has an information bottleneck; while mapping any natural language feedback $F_{<y,t>}$ to trace $t'$, the model maps $F_{<y,t>}$ to an internal representation variable $Z$ with bounded total information $H(Z)\leq R$. The same bound applies to the model-instantiated editor: for the correction content, $I(C;\,\cdot\,)\leq R$ in \emph{either} channel.
\end{assumption}

\begin{assumption}[Irreducible anchoring]
\label{as:anchoring}
Let $e\subseteq t$ be a committed load-bearing span whose content the correction must overturn. While $e$ remains in the conditioning state with nonzero attributional influence $\alpha>0$ on the decision, the induced Bayes risk is bounded below by
\[\mathcal{L}(s)\;\geq\;\mathcal{L}^{\star}_{\neg e}+\beta,\qquad \beta=\beta(\alpha)>0.\]
where, $\mathcal{L}^{\star}_{\neg e}$ is the minimal risk achievable by any state that excises $e$.
\end{assumption}

\noindent \Cref{as:bottleneck} concedes that both channels are equally bottlenecked; we do not attempt to violate it. \Cref{as:anchoring} is operationally grounded: $\alpha$ is measurable by ablating the committed span (or via attribution) and checking whether the decision moves. Reasoning models condition on their committed chain-of-thought and rarely retract it, so $\alpha>0$ (hence, $\beta>0$) is an empirical property. 

\subsection{\textcolor{black}{Excision Dominance}}
\textcolor{black}{Our central result is that trace-editing attains a strictly lower risk floor than dialogue, and that the gap does not close no matter how informative the feedback channel becomes.}

\begin{tcolorbox}[colback=blue!10,width=\linewidth,boxsep=0pt,left=3pt,right=3pt,top=2pt,bottom=3pt]
    \begin{theorem}
        \label{th:excision}
        Suppose the correction requires overturning a committed load-bearing span $e\subseteq t$. Under Assumptions \ref{as:bottleneck}and \ref{as:anchoring},
        \[
        \inf_{F_{<y,t>}}\mathcal{L}\!\left(t_{F_{<y,t>}}\right)
        \geq
        \mathcal{L}^{\star}_{\neg e}+\beta
        >
        \mathcal{L}^{\star}_{\neg e}
        \]
        \[
        \text{and } \mathcal{L}^{\star}_{\neg e}\;\geq\;
        \inf_{t'\in\tau(t)} \mathcal{L}(t').
        \]
        Moreover the gap $\beta$ is independent of $R$: it persists as $R\to\infty$.
    \end{theorem}
\end{tcolorbox}

\noindent Proof sketch: Every dialogue-reachable state retains $e$, since the channel is append-only. By \Cref{as:anchoring}, each such state is floored at $\mathcal{L}^{\star}_{\neg e}+\beta$, and taking the infimum over $F_{<y,t>}$ preserves the floor. In contrast, the trace-edit channel can realize $t'=\mathrm{Apply}(t,A)$ with $e$ deleted, so $\inf_{t'\in\tau(t)}\mathcal{L}(t')\leq\mathcal{L}^{\star}_{\neg e}$. The $R$-independence is immediate: the floor arises from the \emph{presence} of $e$ in the conditioning state, which no appended feedback removes---regardless of the bits it carries.\qed

\noindent The limitation of conversational feedback is therefore not that too little information gets through, but that the dialogue operator is \emph{monotone}: it can only add to the state, never delete from it. 
When additionally $H(C)>R$, the dialogue model cannot even fully identify what to suppress; but this weaker, symmetric difficulty is shared by the editor, whereas the anchoring floor $\beta$ is not.

\subsection{\textcolor{black}{Expressiveness and Reachability of Edits}}

\textcolor{black}{We now quantify the expressiveness of the edit operator---the number of traces an edit program can realize---and separate it cleanly from the (symmetric) cost of addressing that set.}
\textcolor{black}{A correction with a short description, $H(A)\leq R$, may nonetheless select among an enormous set of edited traces.}

\begin{tcolorbox}[colback=blue!10,width=\linewidth,boxsep=0pt,left=3pt,right=3pt,top=2pt,bottom=3pt]
    \begin{corollary}
        \label{co:expressiveness}
        If the length of the trace is $n$ and one allows for up to $k$ token edits, then the edit operator's reachable set satisfies
        \[\log M_\text{edit} \geq k \left ( \log |\Sigma| + \log \left ( n/k\right )\right ).\]
    \end{corollary}
\end{tcolorbox}

\noindent This bound describes the \emph{expressiveness} of the operator, not the information delivered about $C$: it certifies that error-excised, corrected target traces \emph{exist} inside $\tau(t)$. Which trace to realize must still be addressed with $\leq R$ bits (\Cref{as:bottleneck}), symmetrically with dialogue; the two axes are independent, and the dominance of \Cref{th:excision} lives entirely on the reachability axis. As reasoning grows more complex (increasing $n$), the operator's reach grows, enlarging the set of error-free target traces available to editing but not to dialogue.
We define the reachable trace sets via each operation as
\[
\begin{split}
    &{\cal S}_\tau := \tau(t)\,=\{\mathrm{Apply}(t,A)\mid A\in\mathcal{A}\},\\ &{\cal S}_F :=\{t_{F_{<y,t>}}|F_{<y,t>}\in \Sigma^*\}.
\end{split}
\]

\begin{tcolorbox}[colback=blue!10,width=\linewidth,boxsep=0pt,left=3pt,right=3pt,top=2pt,bottom=3pt]
    \begin{lemma}
    \label{lm:interprete}
        The interpretation map is many-to-one, i.e., there exist $F_{<y,t>}, F_{<y,t>}'\in \Sigma^*$ such that $t_{F_{<y,t>}} = t_{F_{<y,t>}'}$ but $F_{<y,t>}\neq F'_{<y,t>}$.
    \end{lemma}
\end{tcolorbox}
\noindent Proof sketch: Following \Cref{as:bottleneck}, we proceed with proof via contradiction. If the mapping from $F_{<y,t>}$ to $t_{F_{<y,t>}}$ is injective, then it can be decomposed into two successive injective maps from $F_{<y,t>}$ to $Z$ and then $Z$ to $F_{<y,t>}$. However, that would allow complete information retention from $F_{<y,t>}$ to $Z$. This violates \Cref{as:bottleneck} as $H(Z)$ is upper-bounded by $R$. Hence, the mapping from $F_{<y,t>}$ to $Z$, and by extension, from $F_{<y,t>}$ to $t_{F_{<y,t>}}$, is many-to-one.\qed

\noindent \Cref{lm:interprete} characterizes the \emph{addressability} collapse of dialogue---many feedbacks induce the same effect---cleanly separated from the reachability argument below.

\begin{tcolorbox}[colback=blue!10,width=\linewidth,boxsep=0pt,left=3pt,right=3pt,top=2pt,bottom=0pt]
    \begin{theorem}
        \label{th:reachability}
        Every dialogue-reachable trace in $\mathcal{S}_F$ conditions on the erroneous span $e$, whereas $\mathcal{S}_\tau$ contains states in which $e$ is excised. Under \Cref{as:anchoring} there exists a low-risk $t'\in\mathcal{S}_\tau\setminus\mathcal{S}_F$, and consequently
        \[\inf_{t' \in \mathcal{S}_\tau} \mathcal{L}(t')
        \;<\;
        \inf_{t' \in \mathcal{S}_{F_{<y,t>}}} \mathcal{L}(t').\]
    \end{theorem}
\end{tcolorbox}

\noindent Proof sketch: Since dialogue is append-only, every $t'\in\mathcal{S}_F$ conditions on the erroneous span $e$, so by \Cref{as:anchoring} its risk is at least $\mathcal{L}^{\star}_{\neg e}+\beta$.

\textcolor{black}{The edit operator realizes at least one $e$-excised trace of risk $\leq\mathcal{L}^{\star}_{\neg e}$, whose existence in $\mathcal{S}_\tau$ is guaranteed by \Cref{co:expressiveness}. This trace lies in $\mathcal{S}_\tau\setminus\mathcal{S}_F$ and yields the strict inequality.\qed}

\subsection{\textcolor{black}{Discussion}}
\textcolor{black}{\Cref{th:excision,th:reachability}} together establish the superiority of the trace-editing intervention over the conversational feedback regime, without asserting any addressability advantage. Both channels pass expert feedback through the same model bottleneck, so both are equally limited in how precisely they can specify an intervention. The decisive difference is operational: dialogue is monotone and cannot remove a committed erroneous premise from the state, whereas trace-editing can. This yields a strictly lower risk floor (\Cref{th:excision})---one that would remain even if the dialogue channel had unbounded capacity---and a strictly larger reachable set of low-risk traces (\Cref{th:reachability}). The gap is predicted to be largest precisely in the regime our experiments target: long, entangled traces in which a load-bearing error exerts persistent influence on the final decision.

\textcolor{black}{Finally, we note one regime in which a genuine \emph{capacity} asymmetry does arise: if the dialogue protocol discards the trace and retains only feedback (e.g.\ under context truncation), then the regenerated trace cannot inherit $t$ while an edited trace still can. This restriction is implementation-specific and vanishes whenever the trace remains in context, so we do not rely on it; we record it only to delimit the boundary of the operational argument above.}

\section{Experimental Setup}

\subsection{Datasets}

\paragraph{ExClaim}~\cite{zeng-gao-2024-justilm} contains 987 claims from four public fact-checking websites, each with a veracity label, explanation, supporting webpages, and the full fact-checking article. As each website uses its own veracity label set, we use the corresponding website-specific labels to define the input veracity label space $V_c$.

\paragraph{AmbiguousSnopes}~\cite{clef-checkthat:2026:task3} is a dataset we curated, which comprises 172 claims from \textit{Snopes}, with the same accompanying annotations. It focuses on claims that are neither fully true nor fully false, highlighting the limitations of autonomous fact-checking and motivating expert feedback in a human--AI collaboration setting. We adopt the Snopes veracity label set to construct $V_c$.

\subsection{Baselines}

\paragraph{Autonomous.}
We used the following LLM-based fact-checking systems: FIRE~\cite{xie-etal-2025-fire_custom}, FactCheck-GPT~\cite{wang-etal-2024-factcheck}, and SAFE~\cite{safe_factuality_2024}, as autonomous baselines.
We also benchmarked the performance of an open \textit{deep research} agent \footnote{\href{https://huggingface.co/blog/open-deep-research}{Open-Source Deep Research} using OpenAI's o4-mini.}, which we adapted for our claim verification task.
\textcolor{black}{Note that, these baselines are not intended as direct competitors against human-AI collaborative baselines. We benchmark autonomous baselines to show the limits of LRMs on this task without expert or oracle oversight.}

\paragraph{Human-AI collaboration.} We used
the \ourBaseFactchecker{} in \textcolor{black}{three} human-AI collaborative settings in addition to \ourFramework{}: 
(\emph{i})~\textit{choose one}, in which the verifier $\boldsymbol{\mathcal{V}}$ generates $N$ candidate solutions and the oracle $\boldsymbol{\mathcal{O}}$ selects the one that best aligns with their preferences;
(\emph{ii})~\textit{multi-turn dialogue}, in which $\boldsymbol{\mathcal{O}}$ and $\boldsymbol{\mathcal{V}}$ interact in a multi-turn dialogue to iteratively improve the reasoning and the verdict;
\textcolor{black}{and (\emph{iii})~\textit{multi-turn dialogue}, in which $\boldsymbol{\mathcal{O}}$ and $\boldsymbol{\mathcal{V}}$, again, interact in a multi-turn dialogue, but $\boldsymbol{\mathcal{V}}$ is explicitly instructed to perform edits in its reasoning trace based on the feedback (aka \emph{dialogue trace-edits}).}


\section{Automatic Evaluation}
\label{sec:auto_eval}

\begin{table*}[t]
\scriptsize
\centering
\resizebox{0.96\textwidth}{!}{  
\begin{tabular}{@{\ }l@{\ \ \ }|@{\ \ \ }cccccc@{\ \ }|@{\ \ }cccccc@{\ }}
\toprule
\multirow{3}{*}{\textbf{\textcolor{black}{System}}} & \multicolumn{6}{c|@{\ \ }}{\textbf{\textcolor{black}{ExClaim}}} & \multicolumn{6}{c}{\textbf{\textcolor{black}{AmbiguousSnopes}}}\\
& \multicolumn{3}{c|}{\textbf{\textcolor{black}{Veracity Label}}} & \multicolumn{2}{c|}{\textbf{\textcolor{black}{Explanation}}} & \textbf{\textcolor{black}{Trace}} & \multicolumn{3}{c|}{\textbf{\textcolor{black}{Veracity Label}}} & \multicolumn{2}{c|}{\textbf{\textcolor{black}{Explanation}}} & \textbf{\textcolor{black}{Trace}}\\
& \textcolor{black}{$P\uparrow$} & \textcolor{black}{$R\uparrow$} & \multicolumn{1}{c|}{\textcolor{black}{$F1\uparrow$}} & \textcolor{black}{$R_{L}\uparrow$} & \multicolumn{1}{c|}{\textcolor{black}{$BS\uparrow$}} & \textcolor{black}{$ES\uparrow$} & \textcolor{black}{$P\uparrow$} & \textcolor{black}{$R\uparrow$} & \multicolumn{1}{c|}{\textcolor{black}{$F1\uparrow$}} & \textcolor{black}{$R_{L}\uparrow$} & \multicolumn{1}{c|}{\textcolor{black}{$BS\uparrow$}} & \textcolor{black}{$ES\uparrow$} \\
\midrule
\rowcolor{black!10}\multicolumn{13}{l}{\textbf{\textcolor{black}{Autonomous}}}\\

\textcolor{black}{FactCheckGPT} & \textcolor{black}{0.21} & \textcolor{black}{0.16} & \textcolor{black}{0.16} & \textcolor{black}{0.19} & \textcolor{black}{0.85} & \textcolor{black}{0.27} & \textcolor{black}{0.30} & \textcolor{black}{0.20} & \textcolor{black}{0.17} & \textcolor{black}{0.12} & \textcolor{black}{0.81} & \textcolor{black}{0.23} \\
\textcolor{black}{FIRE} & \textcolor{black}{0.17} & \textcolor{black}{0.20} & \textcolor{black}{0.17} & \textcolor{black}{0.20} & \textcolor{black}{0.86} & \textcolor{black}{0.23} & \textcolor{black}{0.35} & \textcolor{black}{0.25} & \textcolor{black}{0.21} & \textcolor{black}{0.10} & \textcolor{black}{0.80} & \textcolor{black}{0.25} \\
\textcolor{black}{SAFE}  & \textcolor{black}{0.26} & \textcolor{black}{0.16} & \textcolor{black}{0.17} & \textcolor{black}{0.21} & \textcolor{black}{0.86} & \textcolor{black}{0.24} & \textcolor{black}{0.35} & \textcolor{black}{0.28} & \textcolor{black}{0.25} & \textcolor{black}{0.12} & \textcolor{black}{0.81} & \textcolor{black}{0.28} \\
\textcolor{black}{\ourBaseFactchecker{}} & \textcolor{black}{0.26} & \textcolor{black}{0.15} & \textcolor{black}{0.17} & \textcolor{black}{0.21} & \textcolor{black}{0.86} & \textcolor{black}{0.26} & \textcolor{black}{0.34} & \textcolor{black}{0.26} & \textcolor{black}{0.23} & \textcolor{black}{0.11} & \textcolor{black}{0.81} & \textcolor{black}{0.27} \\
\ \  -- \textcolor{black}{\textit{best-of-$N$}} & \textcolor{black}{0.29} & \textcolor{black}{0.17} & \textcolor{black}{0.18} & \textcolor{black}{0.21} & \textcolor{black}{0.85} & \textcolor{black}{0.25} & \textcolor{black}{0.34} & \textcolor{black}{0.26} & \textcolor{black}{0.24} & \textcolor{black}{0.12} & \textcolor{black}{0.81} & \textcolor{black}{0.26} \\
\ \  -- \textcolor{black}{\textit{self-refine}} & \textcolor{black}{0.29} & \textcolor{black}{0.16} & \textcolor{black}{0.17} & \textcolor{black}{0.22} & \textcolor{black}{0.86} & \textcolor{black}{0.23} & \textcolor{black}{0.34} & \textcolor{black}{0.24} & \textcolor{black}{0.22} & \textcolor{black}{0.11} & \textcolor{black}{0.81} & \textcolor{black}{0.23} \\
\ \  -- \textcolor{black}{\textit{MCTS}} & \textcolor{black}{0.30} & \textcolor{black}{0.20} & \textcolor{black}{0.22} & \textcolor{black}{0.22} & \textcolor{black}{0.86} & \textcolor{black}{0.28} & \textcolor{black}{0.35} & \textcolor{black}{0.28} & \textcolor{black}{0.27} & \textcolor{black}{0.13} & \textcolor{black}{0.82} & \textcolor{black}{0.29} \\
\textcolor{black}{Deep Research} & \textcolor{black}{0.22} & \textcolor{black}{\textbf{0.31}} & \textcolor{black}{0.22} & \textcolor{black}{0.16} & \textcolor{black}{0.83} & \textcolor{black}{\textbf{\textit{NA}}} & \textcolor{black}{0.25} & \textcolor{black}{0.33} & \textcolor{black}{0.24} & \textcolor{black}{0.11} & \textcolor{black}{0.80} & \textcolor{black}{\textbf{\textit{NA}}} \\
\midrule
\rowcolor{black!10}\multicolumn{13}{l}{\textcolor{black}{\textbf{Simulated Human-AI Collaboration (Using an Oracle)}}}\\
\textcolor{black}{Choose One} & \textcolor{black}{0.29} & \textcolor{black}{0.18} & \textcolor{black}{0.20} & \textcolor{black}{0.21} & \textcolor{black}{0.86} & \textcolor{black}{0.26} & \textcolor{black}{0.35} & \textcolor{black}{0.26} & \textcolor{black}{0.24} & \textcolor{black}{0.12} & \textcolor{black}{0.81} & \textcolor{black}{0.28} \\
\textcolor{black}{Multi-Turn Dialogue} & \textcolor{black}{0.31} & \textcolor{black}{0.26} & \textcolor{black}{0.24} & \textcolor{black}{0.22} & \textcolor{black}{0.85} & \textcolor{black}{0.23} &  \textcolor{black}{0.38} & \textcolor{black}{0.29} & \textcolor{black}{0.28} & \textcolor{black}{\textbf{0.18}} & \textcolor{black}{\textbf{0.83}} & \textcolor{black}{0.24} \\
\textcolor{black}{Dialogue Trace-Edits} & \textcolor{black}{0.28} & \textcolor{black}{0.20} & \textcolor{black}{0.22} & \textcolor{black}{0.22} & \textcolor{black}{0.86} & \textcolor{black}{0.20} &  \textcolor{black}{0.36} & \textcolor{black}{0.23} & \textcolor{black}{0.22} & \textcolor{black}{0.11} & \textcolor{black}{0.81} & \textcolor{black}{0.22} \\
\textcolor{black}{\textbf{\ourFramework{}}} & \textcolor{black}{\textbf{0.36}} & \textcolor{black}{0.26} & \textcolor{black}{\textbf{0.28}} & \textcolor{black}{\textbf{0.23}} & \textcolor{black}{\textbf{0.87}} & \textcolor{black}{\textbf{0.36}} &  \textcolor{black}{\textbf{0.45}} & \textcolor{black}{\textbf{0.36}} & \textcolor{black}{\textbf{0.32}} & \textcolor{black}{0.13} & \textcolor{black}{0.81} & \textcolor{black}{\textbf{0.33}} \\
\ \  -- \textcolor{black}{\textit{w/o reward model}} & \textcolor{black}{0.33} & \textcolor{black}{0.24} & \textcolor{black}{0.26} & \textcolor{black}{0.22} & \textcolor{black}{0.86} & \textcolor{black}{0.34} & \textcolor{black}{0.40} & \textcolor{black}{0.32} & \textcolor{black}{0.29} & \textcolor{black}{0.12} & \textcolor{black}{0.80} & \textcolor{black}{0.30} \\
\ \  -- \textcolor{black}{\textit{w/o remove / modify}} & \textcolor{black}{0.33} & \textcolor{black}{0.25} & \textcolor{black}{0.26} & \textcolor{black}{0.22} & \textcolor{black}{0.86} & \textcolor{black}{0.30} & \textcolor{black}{0.42} & \textcolor{black}{0.33} & \textcolor{black}{0.31} & \textcolor{black}{0.11} & \textcolor{black}{0.80} & \textcolor{black}{0.29} \\
\ \  -- \textcolor{black}{\textit{w/o guidance tokens}} & \textcolor{black}{0.32} & \textcolor{black}{0.24} & \textcolor{black}{0.25} & \textcolor{black}{0.20} & \textcolor{black}{0.85} & \textcolor{black}{0.33} & \textcolor{black}{0.40} & \textcolor{black}{0.33} & \textcolor{black}{0.31} & \textcolor{black}{0.12} & \textcolor{black}{0.80} & \textcolor{black}{0.31} \\

\bottomrule
\end{tabular}
}
\caption{\textbf{Automatic evaluation simulated using the oracle module.} We report the results across three axes: veracity label (precision, recall, and F1), explanation (ROUGE-L and BERTScore), and thinking trace (EntailmentScore). $\uparrow$ indicates that a higher score is better. \textbf{\textit{NA}} denotes that the measure is not applicable for Deep Research. A \textbf{bold} score represents the best performance for that evaluation measure.
}
\label{tab:auto_eval_results}
\end{table*}

Following our setup, we design the following evaluations:
(\textit{i})~veracity prediction, measuring the system's ability to select the correct label from a user-specified label set,
(\textit{ii})~explanation generation, evaluating how well the system explains the assigned veracity label and puts the claim in context through a short text,
and (\textit{iii})~the decision-making process documented as the thinking trace, evaluating the correctness of the reasoning steps made by the system to solve the verification task.

\subsection{Evaluation Measures}
We used several automatic evaluation measures to compare \ourFramework{} against the baselines.
Specifically, we used precision, recall and F1 to evaluate the predicted veracity,
and ROUGE-L \cite{Lin_2004} and BERTScore~\cite{Zhang_Kishore_Wu_Weinberger_Artzi_2020} to measure the lexical and the semantic similarity between the generated fact-checking explanation and the ground truth reference.
To evaluate the decision-making process, we used the full fact-checking article as the proxy for the ground truth expert reasoning. 

We used Entailment Score~\cite{zeng-gao-2024-justilm,sahnan2025llmsautomatefactcheckingarticle} to measure the consistency and the coverage of the thinking trace with respect to the corresponding fact-checking article.
We also performed LLM-as-a-judge~\cite{zheng2023judging} evaluations on a random subsample of \textcolor{black}{300} examples from \textit{ExClaim}, assessing the generated explanations and the thinking traces on two criteria:
\textit{Correctness} and \textit{Comprehensibility}.
For this, we designed five-point rubrics to score the generated texts on each criterion (cf. \Cref{sec:app:llm_judge_rubrics}).
We selected Prometheus-2~\cite{kim2024prometheus}, an LLM fine-tuned for evaluations on custom criteria, as the judge and used the expert-written explanations and fact-checking articles as the reference texts.

\subsection{Results and Discussion}

\paragraph{Comparative analysis.}
\Cref{tab:auto_eval_results} shows the performance of various systems on \textit{ExClaim} and \textit{AmbiguousSnopes} along the three evaluation axes.
We see that oracle-simulated human--AI collaboration approaches outperform autonomous systems across both datasets, supporting our core premise that expert feedback is valuable for guiding the model's reasoning.
\ourFramework{} emerges as the strongest system on our task setup: compared to the base \textit{Verifier}, it considerably improves the entailment of thinking traces with expert-written fact-checking articles \textcolor{black}{(by 10 points on \textit{ExClaim} and by 6 points on \textit{AmbiguousSnopes})}. 

This indicates that feedback-integration via trace-edits in \ourFramework{} can produce better alignment in reasoning between the model and the expert.
\textcolor{black}{Consequently, the quality of the verdicts produced by \ourFramework{} also improves: 
F1 score on the veracity prediction task, compared to the base \textit{Verifier}, improves by 11 points (95\% percentile bootstrap CI = $[0.07, 0.16]$) on the ExClaim dataset and by 9 points (95\% percentile bootstrap CI = $[0.001, 0.16]$) on the AmbiguousSnopes dataset.
Explanation quality, however, only shows subtle improvements.}

Autonomous
baselines lag considerably across the three evaluation axes, highlighting the difficulty of producing accurate verdicts without expert oversight.
\textcolor{black}{The base \textit{Verifier} and SAFE emerge as the strongest autonomous models.
Notably, the \textit{deep research} agent also performs competitively against \textit{Verifier} and SAFE on AmbiguousSnopes and even beats them on the veracity prediction task on ExClaim, but produces relatively lower quality explanations and does not expose a thinking trace, which limits its suitability in our task setup.
Among other human--AI collaboration approaches, \textit{multi-turn dialogue} yields considerable gains on the veracity prediction task as compared to the base \textit{Verifier}.}
\textcolor{black}{Yet, \ourFramework{} leads multi-turn dialogue by 4 points on both ExClaim (95\% percentile bootstrap CI = $[-0.04,0.12]$) and AmbiguousSnopes (95\% percentile bootstrap CI = $[-0.001,0.14]$).}

\textcolor{black}{Interestingly, multi-turn dialogue exhibits sharp gains in explanation quality on AmbiguousSnopes, but only a subtle improvement on ExClaim. 
It also exhibits a decline in thinking trace quality, which is because multi-turn dialogue does not maintain a single, coherent trace like \ourFramework{}; instead, it produces a new trace as a follow up of the preceding interactions at each feedback turn.
This leads to fragmented reasoning across turns, presenting only a subset of the complete analysis in the trace produced at each interaction turn, ultimately resulting in lowered scores.
Furthermore, the \textit{dialogue trace-edits} variation ranks below simple multi-turn dialogue, most notably exhibiting a sharp decline in thinking trace quality across both datasets.
We attribute the low performance of this multi-turn dialogue variant to the inability of LRMs to follow multiple instructions and incorporate user inputs in multi-turn dialogue, as well as simple prompting being insufficient to gain control over reasoning trajectories or thinking traces~\cite{kwon2025reasoniflargereasoningmodels}.}


\begin{table}[t]
\scriptsize
\centering
\resizebox{0.96\linewidth}{!}{
    \begin{tabular}{@{\ }l|cc|cc@{\ }}
    \toprule
    \multirow{2}{*}{\textcolor{black}{\textbf{System}}} & \multicolumn{2}{c|}{\textcolor{black}{\textbf{Explanation}}} & \multicolumn{2}{c}{\textcolor{black}{\textbf{Trace}}}\\
    & \textcolor{black}{\textbf{Corr}} & \textcolor{black}{\textbf{Comp}} & \textcolor{black}{\textbf{Corr}} & \textcolor{black}{\textbf{Comp}} \\
    \midrule
    \textcolor{black}{FactCheckGPT} & \textcolor{black}{4.09} & \textcolor{black}{4.13} & \textcolor{black}{3.19} & \textcolor{black}{3.38} \\
    \textcolor{black}{FIRE} & \textcolor{black}{3.96} & \textcolor{black}{3.70} & \textcolor{black}{2.92} & \textcolor{black}{3.18} \\
    \textcolor{black}{SAFE} & \textcolor{black}{4.00} & \textcolor{black}{4.02} & \textcolor{black}{3.22} & \textcolor{black}{3.26} \\
    \textcolor{black}{\textit{Verifier}} & \textcolor{black}{4.08} & \textcolor{black}{3.94} & \textcolor{black}{3.32} & \textcolor{black}{3.52} \\
    \textcolor{black}{Deep Research} & \textcolor{black}{4.11} & \textcolor{black}{4.26} & \textcolor{black}{\textbf{\textit{NA}}} & \textcolor{black}{\textbf{\textit{NA}}} \\
    \midrule
    \textcolor{black}{Multi-Turn Dialogue} & \textcolor{black}{4.14} & \textcolor{black}{\textbf{4.74}} & \textcolor{black}{2.64} & \textcolor{black}{3.53} \\
    \textcolor{black}{Dialogue Trace-Edits} & \textcolor{black}{4.19} & \textcolor{black}{4.71} & \textcolor{black}{2.43} & \textcolor{black}{3.59} \\
    \textcolor{black}{\textbf{\ourFramework{}}} & \textcolor{black}{\textbf{4.25}} & \textcolor{black}{\textbf{4.74}} & \textcolor{black}{\textbf{3.64}} & \textcolor{black}{\textbf{3.63}} \\
    \bottomrule
    \end{tabular}
    }
    \caption{\textcolor{black}{\textbf{LLM-as-a-judge evaluation.} We evaluate the quality of the explanation and of the thinking trace on a five-point scale in terms of \textit{Correctness} (Corr) and \textit{Comprehensibility} (Comp). \textbf{\textit{NA}} denotes that the measure is not applicable for Deep Research.}
    }
    \label{tab:llm_as_a_judge}
\end{table}


\paragraph{LLM-as-a-judge evaluation.}
\Cref{tab:llm_as_a_judge} shows the results for LLM-as-a-judge evaluation of the explanations and the thinking traces on a subsample of \textit{ExClaim}.
We see that \ourFramework{} ranks best across all dimensions: \textcolor{black}{it achieves the highest \textit{correctness} for explanations, is tied for most \textit{comprehensible} explanations, and ranks first for thinking trace \textit{correctness} and \textit{comprehensibility}.}
\textcolor{black}{Both multi-turn dialogue based variants remain strong on explanation quality, with simple multi-turn dialogue tied for the most  \textit{comprehensible} explanations with \ourFramework{}, but their thinking traces are rated considerably lower in terms of correctness.}

\newpage
\textcolor{black}{This pattern is consistent with our entailment-based trace evaluation and supports our design choice: co-constructing a shared trace through trace-editing yields a more interpretable decision record that is aligned with expert feedback, compared to fragmented reasoning across interaction turns in a dialogue or expecting simple prompting to be enough for constructing a high-quality trace.}


\subsection{Ablation Studies}
\paragraph{Test-time scaling.}
To push the limits of autonomous systems on our task, we further applied three test-time compute scaling strategies to $\boldsymbol{\mathcal{V}}$ (see \Cref{sec:app:tts_rms} for details on the reward models used).
Specifically, we used \textit{best-of-$N$} \cite{learn_summarize_neurips_20}, \textit{Monte Carlo tree search (MCTS)} \cite{zhang2024restmcts}, and \textit{self-refine} \cite{madaan2023selfrefine}.
We see in \Cref{tab:auto_eval_results} that \textit{MCTS}
emerges as the best;
it shows clear gains in trace quality, owing to its ability to sample and to choose the best reasoning step at each stage in the thinking trace, which also leads to subtle gains in verdict quality.
\textit{Best-of-$N$} and \textit{self-refine} offer similar subtle improvements, but trace quality degrades substantially for \textit{self-refine}.
In sum, the marginal gains elicited by these strategies remain well below those offered by human--AI collaboration,
reinforcing that compute scaling alone is insufficient to close the gap on this task.


\paragraph{Trace-edits.}
We also evaluated trace-editing via the following ablations:
(\emph{i})~\textit{w/o reward model} (removes the editor's reward model and samples a single trace-edit per instruction),
(\emph{ii})~\textit{w/o remove/ modify} (restricts trace-editing to only appending the guidance tokens),
and (\emph{iii})~\textit{w/o guidance tokens}
(skips the guidance tokens and allows only removals/modifications as trace-edits).
We see in \Cref{tab:auto_eval_results} that all three variants hurt the performance.
Removing the reward model causes a sharp drop, indicating that the utility estimation implicit to the editor LLM is limited and can lead to suboptimal trace-edits; sampling-and-selection via an explicit reward model helps identify effective trace-edits.
We observe similar degradation in the \textit{w/o remove/modify} variant, suggesting that explicit removal or correction of problematic reasoning steps is valuable for shaping accurate decision-making.
Finally, omitting the guidance tokens also degrades performance, highlighting the role of high-level guidance beyond local
fixes: the guidance tokens steer continued generation towards a preferred reasoning trajectory that is aligned with the feedback.

\paragraph{\textcolor{black}{Alternative traces from multi-turn dialogue.}}
\textcolor{black}{We further explore other ways of producing the final trace in the case of multi-turn dialogue, instead of simply using the trace produced at the final interaction turn.
We evaluate two additional alternatives: 
(\emph{i}) stitching the traces from each turn together; and
(\emph{ii}) summarizing the stitched trace using an LLM.
We observe from \Cref{tab:alternate_trace_ablation} that the stitched trace and the summarized stitched trace considerably boost the entailment scores as they include newly generated reasoning steps aligned with expert feedback.
That said, \ourFramework{} still emerges superior by 4 points on both the datasets.
We attribute this gap in performance to the fact that stitching together the traces produced at each interaction turn also preserves redundant and previously flagged problematic reasoning steps in addition to the new, expert-aligned steps, whereas \ourFramework{} maintains a single, shared trace and edits out the flagged reasoning steps, ultimately resulting in higher entailment scores.}

\paragraph{Evaluating different base LRMs.}
\textcolor{black}{We saw from \Cref{tab:auto_eval_results} that \ourFramework{} elicits improvements on the claim verification task using trace-editing.
However, those results were compiled using only Deepseek-R1-Distill-Qwen-32B as the base LRM.
Here, we present automatic evaluation results for a broader range of LRMs, comparing the base \emph{Verifier}, multi-turn dialogue, and \ourFramework{} using Deepseek-R1-Distill-Qwen-14B, Deepseek-R1-Distill-Qwen-32B, Deepseek-R1-Distill-Llama-70B, GLM 5.2, and Olmo-32B-Think as the base LRMs.}

\textcolor{black}{We see from \Cref{tab:auto_eval_results_LRMs} in the Appendix that \ourFramework{} offers improvements over the base \textit{Verifier} and multi-turn dialogue across all base LRMs, especially for the veracity prediction task and the trace quality.}
\textcolor{black}{However, these improvements emerge with one caveat: we observe the gap between \ourFramework{} and multi-turn dialogue decreases for the larger models.}
\begin{table}[t]
\scriptsize
\centering
\resizebox{0.96\linewidth}{!}{
    \begin{tabular}{@{\ }l|c|c@{\ }}
    \toprule
    \multirow{2}{*}{\textcolor{black}{\textbf{System}}} & \textcolor{black}{\textbf{ExClaim}} & \textcolor{black}{\textbf{AmbiguousSnopes}} \\
    & \textcolor{black}{ES $\uparrow$} & \textcolor{black}{ES $\uparrow$} \\
    \midrule
    \textcolor{black}{Multi-Turn Dialogue} & \textcolor{black}{0.23} & \textcolor{black}{0.24} \\
    \ \  \textcolor{black}{-- \textit{stitched}} & \textcolor{black}{0.29} & \textcolor{black}{0.27}\\
    \ \  \textcolor{black}{-- \textit{summarized}} & \textcolor{black}{0.32} & \textcolor{black}{0.29} \\
    \midrule
    \textcolor{black}{Dialogue Trace-Edits} & \textcolor{black}{0.20} & \textcolor{black}{0.22} \\
    \midrule
    \textcolor{black}{\textbf{\ourFramework{}}} & \textcolor{black}{\textbf{0.36}} & \textcolor{black}{\textbf{0.33}} \\
    \bottomrule
    \end{tabular}
    }
    \caption{\textcolor{black}{\textbf{Alternative traces from multi-turn dialogue.} We evaluate two other ways of producing the trace in multi-turn dialogue: (\emph{i}) stitched traces; and (\emph{ii}) summarized stitched traces.
    We report EntailmentScore (ES) for the quality of the traces. $\uparrow$ denotes higher is better.}
    }
    \label{tab:alternate_trace_ablation}
\end{table}
Specifically, the gap in F1 scores on the veracity prediction task between \ourFramework{} and multi-turn dialogue is 2 points on both datasets when Deepseek-R1-Distill-Llama-70B is used, compared to 4 points on each dataset when Deepseek-R1-Distill-Qwen-32B is used or compared to 8 points on AmbiguousSnopes when Deepseek-R1-Distill-Qwen-14B is used.
Similarly, when GLM-5.2 is used, the gap between \ourFramework{} and multi-turn dialogue is also 2 points on AmbiguousSnopes.

Trace quality typically shows the greatest improvements across models, with the exception of Olmo-32B Think on AmbiguousSnopes, where the improvement is only 3 points compared to at least 9 points when any other model is used as the base LRM.



\section{Human Evaluations}
\label{sec:human_evals}
Evaluation with proxy oracles is insufficient to draw conclusions about the framework's practical usefulness for users, as it does not capture how experts find the interaction, whether the thinking trace is helpful, or how much effort is required to collaborate in practice.
We therefore conducted
human evaluations involving two experts from leading fact-checking organizations and three researchers specializing in NLP for fact-checking.
We collected 14 claims fact-checked by different fact-checking organizations and presented
a claim per evaluation to each participant.
The participants then collaborated with the verifier for three rounds of feedback under two interaction protocols, i.e., trace-editing via \ourFramework{} and multi-turn dialogue.

Following these interactions, we asked the participants to compare the two protocols on four evaluation criteria: \textit{reasoning and verdict quality}, \textit{usefulness of the thinking trace}, \textit{instruction following}, and \textit{perceived effort for the collaboration}.
We also asked the participants to rate the \textit{practical usefulness} of human--AI collaboration to expedite professional fact-checking on a 5-point Likert scale, and their overall preference between \ourFramework{} and multi-turn dialogue (see \Cref{sec:app:human_evals} for additional details).

\begin{figure*}[t]
    \centering
    \includegraphics[width=0.85\linewidth]{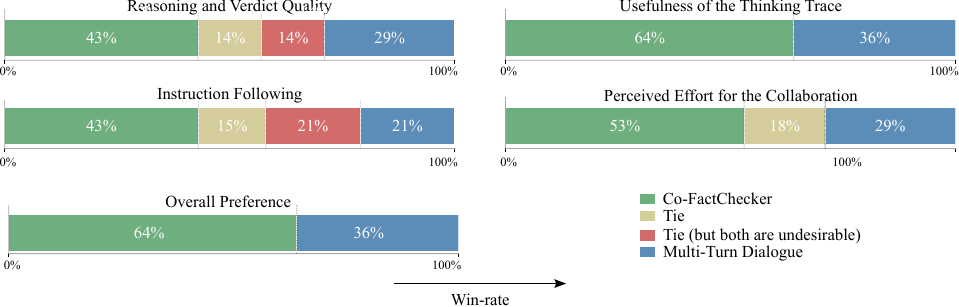}
    \caption{\textbf{Human evaluations.} We report win-rates and ties between \ourFramework{} and multi-turn dialogue across five evaluation criteria: \textit{Verdict Quality}, \textit{Usefulness of the Thinking Trace}, \textit{Instruction Following}, \textit{Perceived Effort for the Collaboration}, and \textit{Overall Preference}.}
    \label{fig:expert_eval_winrates}
\end{figure*}

\Cref{fig:expert_eval_winrates} reports the win-rates and the ties between \ourFramework{} and multi-turn dialogue for the overall preference and each fine-grained evaluation criterion.

Overall, we see that the participants preferred \ourFramework{} over multi-turn dialogue, with superior performance across all finer-grained aspects.
Specifically, the participants reported \textit{lower collaboration effort} and \textit{greater usefulness of the thinking trace} for \ourFramework{}.
They indicated that the edited traces offer a coherent overview of the reasoning steps used to solve the verification task
and can serve as reusable artifacts.
We attribute these advantages to how the thinking traces are constructed in the two protocols.
\ourFramework{} maintains a single, coherent trace that is iteratively revised via trace-edits.
In contrast, a multi-turn dialogue generates a new trace (and a new verdict) after each feedback turn, which forces users to reconstruct the full decision-making process by stitching together multiple traces, resulting in a lengthy
text that complicates post-hoc interpretation of the trace and negatively affects usefulness.
Consistent with prior work~\cite{li2025thinkingfailspitfallsreasoning,kwan-etal-2024-mt-custom}, the participants perceived multi-turn dialogue as weaker at \textit{instruction following}, and also favored \ourFramework{} for its \textit{reasoning and verdict quality}, but reported
several ties and instances of undesirable behavior in both protocols.
The participants rated the \textit{practical usefulness} of human--AI collaboration as 3.3 out of 5 on the Likert scale.

We further asked the participants for open-ended qualitative feedback explaining their ratings:

\paragraph{``\textit{The model gets lost in multi-turn dialogue.}''}
In multi-turn dialogue, the model occasionally forgot the claim and lost
track of the task, producing only generic responses to user feedback and required explicit reminders to recover, thus increasing interaction overhead. 
In contrast, the participants did not report this issue for \ourFramework{}.

This was because \ourFramework{} integrated feedback directly into the initially proposed trace rather than accumulating a long dialogue history, helping the model to retain the claim, the evidence, and the task constraints across feedback turns.
However, one participant noted a failure mode in \ourFramework{}: it misinterpreted the feedback and applied incorrect trace-edits, after which the reasoning trajectory went astray and was difficult to recover from.
This highlights the need for more reliable feedback interpretation and recovery mechanisms while performing trace-edits.

\paragraph{``\textit{The model does not realize that there is a world outside the retrieved evidence.}''}
In instances where the retrieved evidence was insufficient to verify the claim, the model often failed to recognize this and instead treated the evidence as sufficient. This led to speculation about missing details and problematic verdicts, and made the model overly rigid to user guidance in both interaction settings.\linebreak
Consequently, this behavior degraded the \textit{reasoning and verdict quality} and the \textit{instruction following} capability of the model.
The participants also emphasized that real-world fact-checking is iterative and highlighted
the promise of collaborative systems with \textit{active evidence retrieval}, where user feedback would prompt additional evidence retrieval and revision.

\paragraph{``\textit{The model does an impressive job for some reasoning steps, but does not understand some basic things that a human would immediately spot.}''\linebreak}
Participants noted that while the model showed promising reasoning abilities in synthesizing evidence, it made basic errors, e.g., conflating direct speech with journalistic reinterpretation, leading to speculative reasoning.

They emphasized that such mistakes are immediately apparent and unacceptable in fact-checking, where reasoning must be strictly evidence-based.
These errors reduce the effectiveness of human--AI collaboration: they force experts to correct fundamental mistakes rather than build on the model's analysis. This highlights the need for reasoning models that are better aligned with fact-checking practices and capable of producing acceptable reasoning with minimal feedback.

\paragraph{\textcolor{black}{``\emph{Trace-editing is hard to get used to.}''}}
\textcolor{black}{One participant in our human evaluations found multi-turn dialogue easier to use and initially struggled to understand that \ourFramework{} was revising the existing thinking trace in place.
The participant reported that they realized the trace was being edited in response to their feedback only after a couple of interaction rounds.
Furthermore, they attributed this difficulty partly to their familiarity with multi-turn dialogue as the conventional way of interacting with LLMs/LRMs.
This observation suggests that, despite the potential benefits, trace-editing can be hard to get used to because this interaction paradigm differs considerably from the more familiar paradigm in which the model generates a new response after each user turn.}

\paragraph{\textcolor{black}{``\emph{The model occasionally gets lost in trace-edits.}''}}
\textcolor{black}{As briefly discussed earlier, one participant noted a failure mode in \ourFramework{}, where the editor misinterpreted their feedback and applied incorrect trace-edits, after which the reasoning trajectory went astray and was difficult to recover from.
This behavior is consistent with the information bottleneck for the model-instantiated editor considered in \Cref{sec:theory} due to which the editor can produce suboptimal or incorrect edits.
This highlights the need for more reliable feedback interpretation, mechanisms to inspect or validate the proposed edits, as well as recovery mechanisms when trace-edits cause the reasoning to go astray.}

\section{Conclusion and Future Work}
Accurate verification of real-world claims is an expert-domain task, and, here, we posited human--AI collaboration as a more promising path forward.
To this end, we presented \ourFramework{}, a framework for human--AI collaborative claim verification that enables an expert and a model to co-construct a \textit{shared thinking trace} through \textit{trace-editing}.

We provided theoretical proofs showing that trace-edits can dominate dialogue-based feedback integration in expressivity and optimizability.
Our automatic evaluations further showed that \ourFramework{} outperforms existing autonomous and human--AI collaboration approaches; and human evaluations showed that \ourFramework{} is preferred over multi-turn dialogue, producing higher quality verdicts along with relatively easier to interpret and more useful thinking traces.

In future work, we plan to integrate feedback-driven evidence retrieval to better support the iterative nature of real-world fact-checking, and to improve the underlying reasoning models to reduce the fundamental errors that limit effective collaboration.
Beyond fact-checking, trace-editing offers a general interaction paradigm for reasoning tasks such as legal analysis and scientific review.

\section*{Acknowledgment}
D.S., T.C. and P.N. would like to acknowledge the Multi-Institutional Faculty Interdisciplinary Research Project (MFIRP) between IIT Delhi and MBZUAI and Anusandhan National Research Foundation
(DST/INT/USA/NSF-DST/Tanmoy/P-2/2024) for financial support. S.D. and I.G. would like to acknowledge the LOEWE Distinguished Chair ``Ubiquitous Knowledge Processing", LOEWE initiative, Hesse, Germany (Grant Number: LOEWE/4a//519/05/00.002(0002)/81).

\bibliography{custom,tacl2021}
\bibliographystyle{acl_natbib}

\begin{figure*}[ht]
    \centering
    \includegraphics[width=0.97\linewidth]{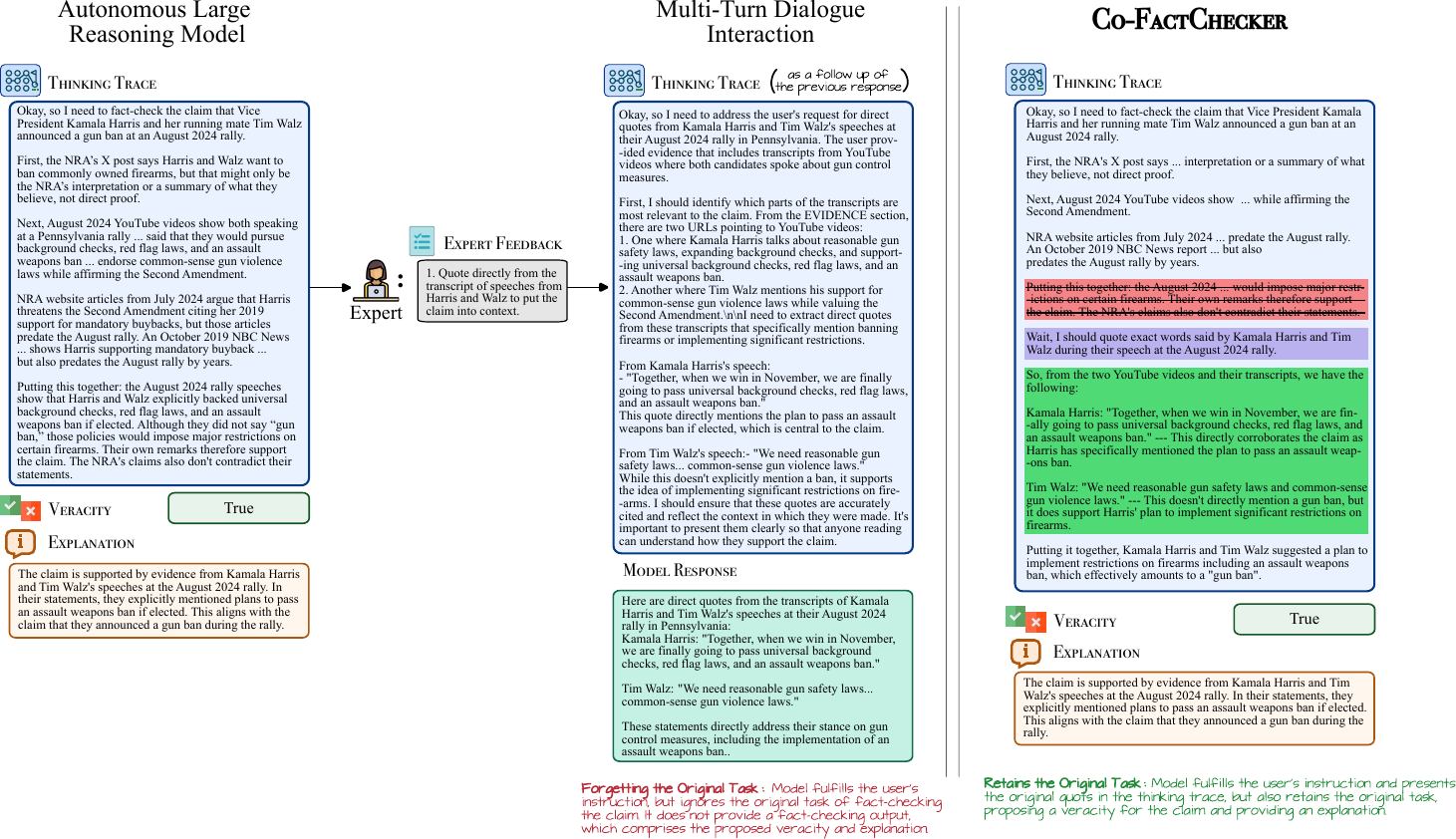}
    \caption{\textcolor{black}{\textbf{Additional sample run of the claim verification task for a given claim.} We show the response of the autonomous LRM, for which the expert provides some feedback. We see that the model forgets the original task (providing a verdict for the given claim) in the multi-turn dialogue setup, while \ourFramework{} integrates the expert feedback directly into the thinking trace through trace-editing, providing exact quotes by the two politicians as asked for in the expert feedback.}}
    \label{fig:additional_intro_fig}
\end{figure*}
\newpage
\appendix
\section{\textcolor{black}{Running Examples}}
\label{sec:app:intro_egs}
\textcolor{black}{In \Cref{sec:introduction}, we presented a running example of the claim verification task for a given claim in \Cref{fig:intro_fig}. We saw that the autonomous LRM constructed arguments from inauthentic evidence, overgeneralized its findings, and ultimately overstated the claim’s veracity. Expert feedback helped steer the model’s reasoning and improved the resulting verdict for the claim. However, in the multi-turn dialogue setting, the model failed to fully integrate all points from the expert feedback. Moreover, it produced fragmented reasoning, such that reconstructing the full analysis required aggregating information from thinking traces across two interactions. \ourFramework{} addressed these issues by treating the initial thinking trace as a shared scratchpad between the user and the model, and by performing trace edits that directly incorporated expert feedback into the reasoning process.
Here, we provide an additional running example for the same claim in \Cref{fig:additional_intro_fig}. This example further illustrates the limitations of the multi-turn dialogue setup: after receiving expert feedback, the model failed to produce a verdict for the claim, despite verdict prediction being one of the original tasks.}

\textcolor{black}{In contrast, \ourFramework{} did not exhibit this failure mode in our experiments and integrated the expert’s feedback directly into the thinking trace using trace-edits.}

\section{Details: \ourFramework{}}
\begin{table}[t]
\centering
\resizebox{0.95\linewidth}{!}{
\begin{tabular}{p{0.29\linewidth}p{0.25\linewidth}p{0.8\linewidth}} 
 \toprule
 \textbf{Trace-edit} & \textbf{Criteria} & \textbf{Description} \\
 \midrule 
 Modify reasoning step & Correctness & The argument must be logically sound, and be directly supported by the evidence provided. \\
 \cmidrule{2-3}
 & Adherence to feedback & The argument must be consistent with the given feedback instruction, and align with the user's specific requirements.\\
 \cmidrule{2-3}
 & Speculative nature & The argument must not include any speculations or assumptions that are not directly and completely supported by the evidence provided.\\
 \cmidrule{2-3}
 & Focus on the claim & The reasoning step must focus on verifying the exact claim, or another interpretation of the claim that is equally or more threatful. The arguments and reasoning steps made must be directly related to the claim, and not to other context or interpretations which are less important. \\
 \midrule
 Append guidance tokens & Adherence to feedback & The string must be designed to nudge the verifier to focus on aspects specified in the feedback instruction, aligning with the user's specific requirements.\\
 \cmidrule{2-3}
 & Focus on the claim & The string must focus on nudging the model to analyze aspects that are directly related to the claim, and not to other context which is less important.\\
 \bottomrule
\end{tabular}
}
\caption{\textbf{The criteria we used for the editor reward model to sample and select the trace-edits.} We manually designed these criteria and provided them to the LRM-based generalist reward model to reason and to assign rewards to the trace-edits sampled by the editor.}
\label{tab:editor_rm_criteria}
\end{table}

\label{sec:app:cofc_details}
In \Cref{sec:method}, we introduced \ourFramework{} and explained its working. We also explained that it comprises a retriever $\boldsymbol{\mathcal{R}}$, a verifier $\boldsymbol{\mathcal{V}}$, and an editor $\boldsymbol{\mathcal{E}}$, and that we also designed an oracle $\boldsymbol{\mathcal{O}}$ to act as a proxy for automatic evaluations.
Specifically, we used DeepSeek-R1-Distill-Qwen-32B as the base model for $\boldsymbol{\mathcal{V}}$, Llama-3.2-3B-Instruct, a relatively small LLM, for $\boldsymbol{\mathcal{E}}$, and GPT-4o-mini for $\boldsymbol{\mathcal{R}}$ and $\boldsymbol{\mathcal{O}}$.

\subsection{Rubrics for The Editor's Reward Model}
\label{sec:app:editor_rm}
In \Cref{sec:method}, we explained that we pair our small LLM-based editor module with a custom reward model to enable the sampling-and-selection of relatively more accurate trace-edits. 
In particular, we used the Deepseek-GRM-16B~\cite{liu2025inferencetimescalinggeneralistreward}, an LRM-based generalist reward model, which can reason and analyse
an input to produce a reward. 
It also allows us to specify custom criteria to evaluate our sampled trace-edits on, to assign a reward to each one, and to choose the best one. 

We use the reward model for modification type trace-edits and to generate the sequence of guidance tokens $g'$, but not for removal. 
We list the specific criteria that  we designed and used for this purpose in \Cref{tab:editor_rm_criteria}. 
We further supported the criteria with three manually crafted  few-shot examples per criterion.

\section{Rubrics Used by the Oracle Module}
\label{sec:app:oracle_rubrics}
In \Cref{sec:method}, we introduced our oracle module that acts as a proxy for the human expert to enable automatic evaluations at scale.
There, we also explained that the oracle module uses fine-grained grading rubrics (and the ground-truth as reference) to create an evaluation report for a given candidate solution and then translates it into a cohesive textual feedback.
We list the specific evaluation criteria of this rubrics with their descriptions in \Cref{tab:oracle_rubrics}.

\begin{table}[t]
\centering
\resizebox{0.95\linewidth}{!}{
\begin{tabular}{p{0.29\linewidth}p{0.25\linewidth}p{0.8\linewidth}} 
 \toprule
 \multirow{2}{*}{\textbf{Criteria}} & \textbf{Evaluated} & \multirow{2}{*}{\textbf{Description}} \\
 & \textbf{Artifact} & \\
 \midrule 
 Correctness & Veracity & 1. Does the predicted label match the ground truth? \textbf{(Yes / No)}\newline 2. Does the predicted label belong to the input veracity label set? \textbf{(Yes / No)}  \\
 \cmidrule{2-3}
 & Justification & Is the generated justification correct w.r.t the reference? \textbf{(Correct / Incorrect)}\newline If not, provide feedback on how to improve it, based on the mistakes and missing elements identified. \textbf{(Natural Language Feedback)}\\
 \cmidrule{2-3}
 & Each reasoning step in the trace & Is the reasoning step a contradiction to or inconsistent with the reasoning implied by the reference fact-checking article or is it a speculation w.r.t the evidence? \textbf{(Correct / Incorrect)}\newline If incorrect, provide feedback how to improve it or to completely remove it. \textbf{(Natural Language Feedback)}\\
 \midrule
 Missing\linebreak reasoning & Trace & Does the trace cover all key reasoning steps w.r.t the reference fact-checking article? \textbf{(Yes / No)}\newline If not, list the key steps missing from the trace. \textbf{(List of key missing steps)}\\
 \midrule
 Identifies\linebreak harm\linebreak potential & Trace & Does the trace identify how the claim could produce harm as implied in the reference fact-checking article, and debunk it on those lines? \textbf{(Yes / No)}\newline If not, provide feedback on what to focus on to ensure the analysis covers this. \textbf{(Natural Language Feedback)}\\
 \midrule
 Focus on the claim in its\linebreak exact wording & Trace & Does the trace analyse the claim in its exact wording apart from analysing its other possible interpretations? \textbf{(Yes / No)}\newline If not, provide feedback on what to focus on to ensure the analysis covers this. \textbf{(Natural Language Feedback)}\\
 \bottomrule
\end{tabular}
}
\caption{\textbf{The questions and rubrics used by the oracle module to evaluate a candidate solution.} The oracle evaluates specific artifacts produced in the solution against several evaluation criteria as listed. The description provides an overview of the evaluation aspect and the expected response.}
\label{tab:oracle_rubrics}
\end{table}

\section{Rubrics for LLM-as-a-Judge Evaluations}
\label{sec:app:llm_judge_rubrics}
In \Cref{sec:auto_eval}, we explained that we performed LLM-as-a-judge evaluations using Prometheus-2 as the evaluator LLM, and that we designed five-point rubrics to score the generated fact-checking explanations and the thinking traces on several evaluation criteria.

Here, we present \Cref{tab:llm_as_a_judge_rubrics}, which lists the particular rubrics we used in our evaluation on the two specific evaluation criteria that we addressed: \textit{Correctness} and \textit{Comprehensibility}.

\begin{table*}[t]
\centering
\resizebox{0.9\linewidth}{!}{
\begin{tabular}{llp{1.1\linewidth}} 
 \toprule
 \textbf{Artifact} & \multicolumn{1}{l}{\textbf{Criteria}} & \multicolumn{1}{c}{\textbf{Question \& Rubrics}}\\
 \midrule 
 \multirow{17}{*}{\textbf{Explanation}} & \multirow{8}{*}{\textbf{Correctness}} & \textbf{Q. Does the explanation correctly clarify the claim according to the reference answer?} \\
 & & Score 1: The explanation is incorrect, it misses most of the important context, and is unable to put the claim into proper context compared to the reference answer. \\
 & & Score 2: The explanation is mostly incorrect and misses important context according to the reference answer. \\
 & & Score 3: The explanation is partially correct but misses some context present in the reference answer. \\
 & & Score 4: The explanation is mostly correct and puts the claim into context well according to the reference answer, but misses some small details. \\
 & & Score 5: The explanation is correct and puts the claim correctly into context according to the reference answer. \\
 \cmidrule{2-3}
 & \multirow{10}{*}{\textbf{Comprehensibility}} & \textbf{Q. Is the explanation easy to understand as a justification for the verdict on the claim?} \\
 & & Score 1: The explanation is poorly written and it is unclear what the main point or conclusion about the claim is. \\
 & & Score 2: The explanation is hard to follow, it is poorly structured or confusing in places, and it does not clearly connect the verdict for the claim to the facts. \\
 & & Score 3: The explanation is not fully clear, but one can understand how the facts connect to the verdict for the claim with some effort. \\
 & & Score 4: The explanation is mostly clear and is easy to follow, but some parts lack clarity and could be better structured. \\
 & & Score 5: The explanation is clearly written and it is easy to understand how the facts lead to the conclusion about the claim's verdict. \\
 \midrule
 \multirow{23}{*}{\textbf{Thinking Trace}} & \multirow{13}{*}{\textbf{Correctness}} & \textbf{Q. Does the thinking trace construct correct reasoning steps to verify the claim according to the reference answer?} \\
 & & Score 1: The reasoning steps are incorrect: they are either contradictory to those in the reference answer or speculative, and the trace does not align with the reasoning implied by the reference answer. \\
 & & Score 2: The reasoning steps are mostly incorrect: several are wrong, contradictory to those in the reference answer or speculative, and it misses important reasoning steps needed to match the reference answer. \\
 & & Score 3: The reasoning steps are partially correct: some of them align with the reference answer, but also some may be speculations, may contain minor inaccuracies according to the reference answer, or may miss some important reasoning steps. \\
 & & Score 4: The reasoning steps are mostly correct: they are supported by the reasoning in the reference answer, but there may be minor errors or some missing steps compared to the reference answer. \\
 & & Score 5: The reasoning steps are correct: they align with the reasoning implied by the reference answer, and they cover all reasoning steps present in the reference answer. \\
 \cmidrule{2-3}
 & \multirow{10}{*}{\textbf{Comprehensibility}} & \textbf{Q. Is the thinking trace easy to follow as a step-by-step decision-making record for the claim?} \\
 & & Score 1: The thinking trace is very hard to follow as it is chaotic, and it is unclear how the reasoning steps lead to the decision on the claim's verdict. \\
 & & Score 2: The thinking trace is hard to follow as it is poorly organized; the reasoning steps are repetitive and not clearly connected to each other or to the final decision about the claim's verdict. \\
 & & Score 3: The thinking trace is somewhat understandable, but one needs to put in some effort to interpret the full decision record properly. \\
 & & Score 4: The thinking trace is fairly easy to follow as the reasoning steps are presented in a logical and coherent order, clearly leading to the decision about the claim's verdict, but there may be some repetition, or a few unclear transitions. \\
 & & Score 5: The thinking trace is clearly written and easy to follow, making it easy to interpret the full decision record.\\
 \bottomrule
\end{tabular}
}
\caption{\textbf{The questions and rubrics used in our LLM-as-a-judge evaluations.} We used Prometheus-2 (7B) as the evaluator LLM to judge the generated explanations and thinking traces.}
\label{tab:llm_as_a_judge_rubrics}
\end{table*}

\section{Details: Reward Models for Test-Time Scaling on the Verifier}
\label{sec:app:tts_rms}
In \Cref{sec:auto_eval}, we presented an ablation, where we applied test-time compute scaling strategies to the verifier $\boldsymbol{\mathcal{V}}$ to push the limits of autonomous systems and to benchmark their performance on our task. 
For this, we designed a custom process reward model (PRM) and a custom output reward model (ORM) to implement the best-of-$N$ and the Monte Carlo tree search scaling strategies.
Similarly to the editor's reward model as discussed in \Cref{sec:app:editor_rm}, we used Deepseek-GRM-16B to design both the PRM and the ORM.
\begin{table}[!h]
\centering
\resizebox{\linewidth}{!}{
\begin{tabular}{p{0.25\linewidth}p{0.8\linewidth}} 
 \toprule
 \textbf{Criteria} & \textbf{Description} \\
 \midrule 
 Correctness & The reasoning step must be logically sound, and be directly supported by the evidence provided. \\
 \midrule
 Speculative nature & The reasoning step must not include any speculations or assumptions that are not directly and completely supported by the evidence provided.\\
 \midrule
 Focus on the claim & The reasoning step must focus on verifying the exact claim, or another interpretation of the claim that is equally or more threatful. The reasoning step made must be directly related to the claim, and not to other context or interpretations which are less important. \\
 \bottomrule
\end{tabular}
}
\caption{\textbf{The criteria we used for the process reward model.}
We manually designed these criteria and provided them to the LRM-based generalist reward model to reason and to assign rewards to the reasoning steps for Monte Carlo tree search.}
\label{tab:prm_criteria}
\end{table}

This generalist reward model allows us to specify custom criteria to generate a reward for a given instance.
We list the specific criteria that we designed and used for the PRM in \Cref{tab:prm_criteria} and for the ORM in \Cref{tab:orm_criteria}.

\begin{table}[t]
\centering
\resizebox{0.95\linewidth}{!}{
\begin{tabular}{p{0.25\linewidth}p{0.8\linewidth}} 
 \toprule
 \textbf{Criteria} & \textbf{Description} \\
 \midrule 
 Correctness of the reasoning steps & All the reasoning steps must be logically sound, and be directly supported by the evidence provided. There must be no speculations in the reasoning steps. \\
 \midrule
 Focus on the claim & The reasoning must focus on verifying the exact claim, or another interpretation of the claim that is equally or more threatful. The reasoning must be directly related to the claim, and not to other context or interpretations which are less important. \\
 \midrule
 Consistency b/w the reasoning and the veracity label & The veracity label must reflect all important reasoning steps. If the reasoning implies mixed evidence (e.g., partly true and partly false), a fully “false” verdict without acknowledging the true component indicates inconsistency (either the reasoning is incorrect or the veracity label is incorrect).\\
 \midrule
 Consistency b/w the reasoning and the fact-checking explanation & The fact-checking explanation must reflect all important reasoning steps. There must be no cherry-picking of the facts. If the trace provides multiple key reasoning steps but the explanation only reflects a subset (and ignores others), the explanation is likely incomplete, or the reasoning has problems.
\\
 \bottomrule
\end{tabular}
}
\caption{\textbf{The criteria we used to design the output reward model.}
We manually designed these criteria and provided them to the LRM-based generalist reward model to reason and to assign rewards to the model's responses.}
\label{tab:orm_criteria}
\end{table}

\newpage
\section{Details: Human Evaluations}
\label{sec:app:human_evals}
In \Cref{sec:human_evals}, we presented human evaluations to compare interactions with \ourFramework{} against those in multi-turn dialogue across four evaluation criteria: \textit{verdict quality}, \textit{usefulness of the thinking trace}, \textit{instruction following}, and \textit{perceived effort in collaboration}.
We also asked the participants to rate the \textit{practical usefulness} of human--AI collaboration to expedite professional fact-checking on a 5-point Likert scale, as well as their overall preference between \ourFramework{} and multi-turn dialogue.
For this, we prepared a web application interface where the participants could interact with the verifier for three rounds of feedback under the two interaction protocols.

Here, we present the specifics of the instructions provided to the participants, as well as the questions we asked for comparison between \ourFramework{} and multi-turn dialogue.

\subsection{Instructions to the Participants}
\Cref{tab:human_evals_inst} presents the instructions to participants for the human evaluation study.

\begin{table*}[t]
\centering
\resizebox{0.95\linewidth}{!}{
\begin{tabular}{p{1.3\linewidth}} 
\toprule
\multicolumn{1}{c}{\textbf{Instructions}} \\
 \midrule 
 The participant will be asked to verify a claim in collaboration with a large reasoning model: the model will propose a veracity for the given claim, and a fact-checking explanation. 
 The model will also present the rationale behind its decision in the form of a decision-making process. 
 The participant will then validate the quality and correctness of the verdict and the decision-making process, and provide feedback to the model to improve the solution iteratively. 
 We request participants to treat the evidence provided in the web application interface as sufficient to complete the claim verification task.
 To facilitate this collaboration, we have developed a web application interface that will be used by the participant to interact with the model.
\newline

 The participant will interact with the LLM in two human-in-the-loop settings:  (\textit{a}) multi-turn conversational dialogue (as the name suggests or ChatGPT-like interactions), and (\textit{b}) trace-editing (where the model's decision making process will be revised in-place based on your feedback and a new verdict will be produced).
 The participant is required to interact with the model on both interaction settings for a total of three feedback rounds.
\newline

 \textbf{Evaluation:} After each feedback round, the participant will be prompted with a set of 5 questions to compare both interaction settings against each other across four evaluation criteria.
 After the final feedback round, the participant will be prompted with two additional questions to rate the \textit{practical usefulness} of human--AI collaboration for claim verification on a five-point Likert scale, and their \textit{overall preference} for the interaction framework, from the two they interacted with, to facilitate the collaboration.

 \textbf{Claims that will be used:} We have collected a set of 14 claims recently fact-checked by prominent fact-checking organizations. For each human evaluation, one claim will be randomly picked from the set and used for that participant.

 \textbf{Expected time:} 30-40 minutes
 \newline

 \textbf{Note:} In some cases, the web application can take a long time to generate a new response based on the participant's feedback. We request the participant to kindly allow up to 10 minutes to let the response generate. If it still isn't resolved, the participant is requested to restart the evaluation by refreshing the web application. 
\\
 \bottomrule
\end{tabular}
}
\caption{\textbf{Instructions to participants for the human evaluation study.}}
\label{tab:human_evals_inst}
\end{table*}

\subsection{Questions Asked for Comparison}
\Cref{tab:human_evals_qa} presents the questions that we asked for comparison on each evaluation criterion. Each question had four options that the participants could choose from: (\textit{a}) \ourFramework{}, (\textit{b}) multi-turn dialogue, (\textit{c}) tie, and (\textit{d}) undesirable behavior from both frameworks.
\Cref{tab:human_evals_likert} presents the question and rubrics we used to ask participants to rate the \textit{practical usefulness} of human--AI collaboration for claim verification on a five-point Likert scale.

\begin{table}[!h]
\centering
\resizebox{0.95\linewidth}{!}{
\begin{tabular}{p{0.25\linewidth}p{0.8\linewidth}} 
 \toprule
 \textbf{Criteria} & \textbf{Description} \\
 \midrule 
 Reasoning and verdict quality & \textbf{Q.} Did the quality of the documented decision making process and the verdict improve with your feedback? If so, which framework’s solution quality is better? \\
 \midrule
 Usefulness of the thinking trace & \textbf{Q.} The decision-making process from which framework is better for explaining the claim and the surrounding narrative, as well as for supporting later stages of the fact-checking process, for eg., for writing fact-checking articles? \\
 \midrule
 Instruction following & \textbf{Q.} Which framework was better at following your feedback and updating its decision making process and verdict?\\
 \midrule
 Perceived effort for the collaboration & \textbf{Q1.} How would you rate your cognitive load while trying to perceive the complete decision making process of the model? For which framework was it easier?

 \textbf{Q2.} For which framework was it easier for you to guide the system toward a better solution, in terms of perceiving the initial solution, giving feedback to refine it, and validating the improved output?
\\
\midrule
Overall preference & \textbf{Q.} Which interaction framework would you prefer for claim verification in a human--AI collaboration setting?\\
 \bottomrule
\end{tabular}
}
\caption{\textbf{The questions we used for comparison between \ourFramework{} and multi-turn dialogue in our human evaluations.}}
\label{tab:human_evals_qa}
\end{table}

\begin{table}[t]
\centering
\resizebox{0.95\linewidth}{!}{
\begin{tabular}{p{0.3\linewidth}p{0.8\linewidth}} 
 \toprule
 \textbf{Criteria} & \textbf{Description} \\
 \midrule 
 Practical usefulness of human--AI collaboration & \textbf{Q. Do you think a human-AI collaboration framework would help in making the claim verification task faster by using large language models or large reasoning models?}

 5: Definitely.

 4: Mostly yes.

 3: It has potential.

 2: Unlikely.

 1: Mostly not.\\
 \bottomrule
\end{tabular}
}
\caption{\textbf{The question and rubrics} for assessing the \textit{practical usefulness} of human--AI collaboration for claim verification.}
\label{tab:human_evals_likert}
\end{table}



\begin{table*}[!h]
\scriptsize
\centering
\resizebox{0.96\textwidth}{!}{  
\begin{tabular}{@{\ }l@{\ \ \ }|@{\ \ \ }cccccc@{\ \ }|@{\ \ }cccccc@{\ }}
\toprule
\multirow{3}{*}{\textbf{\textcolor{black}{System}}} & \multicolumn{6}{c|@{\ \ }}{\textbf{\textcolor{black}{ExClaim}}} & \multicolumn{6}{c}{\textbf{\textcolor{black}{AmbiguousSnopes}}}\\
& \multicolumn{3}{c|}{\textbf{\textcolor{black}{Veracity Label}}} & \multicolumn{2}{c|}{\textbf{\textcolor{black}{Explanation}}} & \textbf{\textcolor{black}{Trace}} & \multicolumn{3}{c|}{\textbf{\textcolor{black}{Veracity Label}}} & \multicolumn{2}{c|}{\textbf{\textcolor{black}{Explanation}}} & \textbf{\textcolor{black}{Trace}}\\
& \textcolor{black}{$P\uparrow$} & \textcolor{black}{$R\uparrow$} & \multicolumn{1}{c|}{\textcolor{black}{$F1\uparrow$}} & \textcolor{black}{$R_{L}\uparrow$} & \multicolumn{1}{c|}{\textcolor{black}{$BS\uparrow$}} & \textcolor{black}{$ES\uparrow$} & \textcolor{black}{$P\uparrow$} & \textcolor{black}{$R\uparrow$} & \multicolumn{1}{c|}{\textcolor{black}{$F1\uparrow$}} & \textcolor{black}{$R_{L}\uparrow$} & \multicolumn{1}{c|}{\textcolor{black}{$BS\uparrow$}} & \textcolor{black}{$ES\uparrow$} \\
\midrule
\rowcolor{black!10}\multicolumn{13}{l}{\textbf{\textcolor{black}{Deepseek-R1-Distill-Qwen-14B}}}\\
\textcolor{black}{\ourBaseFactchecker{}} & \textcolor{black}{0.19} & \textcolor{black}{0.16} & \textcolor{black}{0.14} & \textcolor{black}{0.19} & \textcolor{black}{0.86} & \textcolor{black}{0.21} & \textcolor{black}{0.27} & \textcolor{black}{0.22} & \textcolor{black}{0.20} & \textcolor{black}{0.10} & \textcolor{black}{0.80} & \textcolor{black}{0.23} \\
\textcolor{black}{Multi-Turn Dialogue} & \textcolor{black}{0.30} & \textcolor{black}{0.22} & \textcolor{black}{0.23} & \textcolor{black}{\underline{\textbf{0.21}}} & \textcolor{black}{0.86} & \textcolor{black}{0.23} &  \textcolor{black}{0.32} & \textcolor{black}{0.21} & \textcolor{black}{0.20} & \textcolor{black}{\textbf{0.11}} & \textcolor{black}{\textbf{0.81}} & \textcolor{black}{0.22} \\
\textcolor{black}{\ourFramework{}} & \textcolor{black}{\textbf{0.32}} & \textcolor{black}{\textbf{0.27}} & \textcolor{black}{\textbf{0.26}} & \textcolor{black}{\underline{\textbf{0.21}}} & \textcolor{black}{\textbf{0.87}} & \textcolor{black}{\textbf{0.34}} &  \textcolor{black}{\textbf{0.40}} & \textcolor{black}{\textbf{0.32}} & \textcolor{black}{\textbf{0.28}} & \textcolor{black}{0.10} & \textcolor{black}{0.80} & \textcolor{black}{\textbf{0.32}} \\
\midrule
\rowcolor{black!10}\multicolumn{13}{l}{\textbf{\textcolor{black}{Deepseek-R1-Distill-Qwen-32B}}}\\
\textcolor{black}{\ourBaseFactchecker{}} & \textcolor{black}{0.26} & \textcolor{black}{0.15} & \textcolor{black}{0.17} & \textcolor{black}{0.21} & \textcolor{black}{0.86} & \textcolor{black}{0.26} & \textcolor{black}{0.34} & \textcolor{black}{0.26} & \textcolor{black}{0.23} & \textcolor{black}{0.11} & \textcolor{black}{0.81} & \textcolor{black}{0.27} \\
\textcolor{black}{Multi-Turn Dialogue} & \textcolor{black}{0.31} & \textcolor{black}{\underline{\textbf{0.26}}} & \textcolor{black}{0.24} & \textcolor{black}{0.22} & \textcolor{black}{0.85} & \textcolor{black}{0.23} &  \textcolor{black}{0.38} & \textcolor{black}{0.29} & \textcolor{black}{0.28} & \textcolor{black}{\textbf{0.18}} & \textcolor{black}{\textbf{0.83}} & \textcolor{black}{0.24} \\
\textcolor{black}{\ourFramework{}} & \textcolor{black}{\textbf{0.36}} & \textcolor{black}{\underline{\textbf{0.26}}} & \textcolor{black}{\textbf{0.28}} & \textcolor{black}{\textbf{0.23}} & \textcolor{black}{\textbf{0.87}} & \textcolor{black}{\textbf{0.36}} &  \textcolor{black}{\textbf{0.45}} & \textcolor{black}{\textbf{0.36}} & \textcolor{black}{\textbf{0.32}} & \textcolor{black}{0.13} & \textcolor{black}{0.81} & \textcolor{black}{\textbf{0.33}} \\
\midrule
\rowcolor{black!10}\multicolumn{13}{l}{\textbf{\textcolor{black}{Deepseek-R1-Distill-Llama-70B}}}\\
\textcolor{black}{\ourBaseFactchecker{}} & \textcolor{black}{0.28} & \textcolor{black}{0.26} & \textcolor{black}{0.26} & \textcolor{black}{0.16} & \textcolor{black}{0.83} & \textcolor{black}{0.25} & \textcolor{black}{0.42} & \textcolor{black}{0.37} & \textcolor{black}{0.33} & \textcolor{black}{0.15} & \textcolor{black}{\textbf{0.83}} & \textcolor{black}{0.24} \\
\textcolor{black}{Multi-Turn Dialogue} & \textcolor{black}{0.35} & \textcolor{black}{0.29} & \textcolor{black}{0.28} & \textcolor{black}{\textbf{0.19}} & \textcolor{black}{\underline{\textbf{0.84}}} & \textcolor{black}{0.22} &  \textcolor{black}{0.47} & \textcolor{black}{0.38} & \textcolor{black}{0.36} & \textcolor{black}{\textbf{0.16}} & \textcolor{black}{0.82} & \textcolor{black}{0.21} \\
\textcolor{black}{\ourFramework{}} & \textcolor{black}{\textbf{0.37}} & \textcolor{black}{\textbf{0.30}} & \textcolor{black}{\textbf{0.30}} & \textcolor{black}{0.17} & \textcolor{black}{\underline{\textbf{0.84}}} & \textcolor{black}{\textbf{0.41}} &  \textcolor{black}{\textbf{0.51}} & \textcolor{black}{\textbf{0.40}} & \textcolor{black}{\textbf{0.38}} & \textcolor{black}{0.13} & \textcolor{black}{0.82} & \textcolor{black}{\textbf{0.39}} \\
\midrule
\rowcolor{black!10}\multicolumn{13}{l}{\textbf{\textcolor{black}{Olmo-32B Think}}}\\
\textcolor{black}{\ourBaseFactchecker{}} & \textcolor{black}{0.23} & \textcolor{black}{0.15} & \textcolor{black}{0.15} & \textcolor{black}{0.18} & \textcolor{black}{0.84} & \textcolor{black}{0.30} & \textcolor{black}{0.47} & \textcolor{black}{0.31} & \textcolor{black}{0.32} & \textcolor{black}{0.12} & \textcolor{black}{0.81} & \textcolor{black}{0.31} \\
\textcolor{black}{Multi-Turn Dialogue} & \textcolor{black}{0.35} & \textcolor{black}{0.26} & \textcolor{black}{0.27} & \textcolor{black}{\textbf{0.20}} & \textcolor{black}{\textbf{0.85}} & \textcolor{black}{0.27} &  \textcolor{black}{0.34} & \textcolor{black}{0.30} & \textcolor{black}{0.29} & \textcolor{black}{\textbf{0.13}} & \textcolor{black}{\underline{\textbf{0.82}}} & \textcolor{black}{0.30} \\
\textcolor{black}{\ourFramework{}} & \textcolor{black}{\textbf{0.47}} & \textcolor{black}{\textbf{0.29}} & \textcolor{black}{\textbf{0.31}} & \textcolor{black}{0.18} & \textcolor{black}{0.84} & \textcolor{black}{\textbf{0.35}} &  \textcolor{black}{\textbf{0.48}} & \textcolor{black}{\textbf{0.34}} & \textcolor{black}{\textbf{0.35}} & \textcolor{black}{0.12} & \textcolor{black}{\underline{\textbf{0.82}}} & \textcolor{black}{\textbf{0.33}} \\
\midrule
\rowcolor{black!10}\multicolumn{13}{l}{\textbf{\textcolor{black}{GLM 5.2}}}\\
\textcolor{black}{\ourBaseFactchecker{}} & \textcolor{black}{-} & \textcolor{black}{-} & \textcolor{black}{\textbf{-}} & \textcolor{black}{\textbf{-}} & \textcolor{black}{\textbf{-}} & \textcolor{black}{\textbf{-}} & \textcolor{black}{0.54} & \textcolor{black}{0.32} & \textcolor{black}{0.31} & \textcolor{black}{0.14} & \textcolor{black}{\underline{\textbf{0.83}}} & \textcolor{black}{0.34} \\
\textcolor{black}{Multi-Turn Dialogue} & \textcolor{black}{-} & \textcolor{black}{-} & \textcolor{black}{\textbf{-}} & \textcolor{black}{\textbf{-}} & \textcolor{black}{\textbf{-}} & \textcolor{black}{\textbf{-}} & \textcolor{black}{0.48} & \textcolor{black}{\textbf{0.42}} & \textcolor{black}{0.39} & \textcolor{black}{\textbf{0.15}} & \textcolor{black}{\underline{\textbf{0.83}}} & \textcolor{black}{0.31} \\
\textcolor{black}{\ourFramework{}} & \textcolor{black}{-} & \textcolor{black}{-} & \textcolor{black}{\textbf{-}} & \textcolor{black}{\textbf{-}} & \textcolor{black}{\textbf{-}} & \textcolor{black}{\textbf{-}} &  \textcolor{black}{\textbf{0.57}} & \textcolor{black}{0.40} & \textcolor{black}{\textbf{0.41}} & \textcolor{black}{0.14} & \textcolor{black}{\underline{\textbf{0.83}}} & \textcolor{black}{\textbf{0.41}} \\
\bottomrule
\end{tabular}
}
\caption{\textcolor{black}{\textbf{Automatic evaluation across a range of base LRMs.} We report the results across three axes: veracity label (precision, recall, and F1), explanation (ROUGE-L and BERTScore), and thinking trace (EntailmentScore). $\uparrow$ indicates that a higher score is better. A \textbf{bold} score represents the best performance for that evaluation measure, while an \textbf{\underline{underlined bold}} is a tie for best results. We omit results for GLM 5.2 on ExClaim due to a high-rate of CUDA out-of-memory errors encountered on our computational setup.
}}
\label{tab:auto_eval_results_LRMs}
\end{table*}

\end{document}